\documentclass[11pt]{article}

\usepackage[]{ACL2023}

\usepackage{times}
\usepackage{latexsym}
\usepackage{tikz}

\usepackage[T1]{fontenc}
\usepackage[utf8]{inputenc}

\usepackage{microtype}

\usepackage{inconsolata}
\usepackage{graphicx}
\usepackage{booktabs}
\usepackage{soul}
\usepackage{multirow}
\usepackage{subfig}
\usepackage{amsmath}
\usepackage{cleveref}
\usepackage{xcolor}
\usepackage{algorithm}
\usepackage{listings}
\usepackage{comment}

\usepackage{amsmath,amsfonts,bm}

\def\eqref#1{equation~\ref{#1}}
\def\1{\bm{1}}

\DeclareMathAlphabet{\mathsfit}{\encodingdefault}{\sfdefault}{m}{sl}
\SetMathAlphabet{\mathsfit}{bold}{\encodingdefault}{\sfdefault}{bx}{n}

\definecolor{myYellow}{rgb}{0.9,0.9,1}
\sethlcolor{myYellow}
\crefformat{section}{\S#2#1#3} 
\crefformat{subsection}{\S#2#1#3}
\crefformat{subsubsection}{\S#2#1#3}

\newenvironment{pcrfont}{\fontfamily{pcr}\selectfont}{\par}
\DeclareTextFontCommand{\textpcr}{\pcrfont}

\definecolor{airforceblue}{rgb}{0.14, 0.31, 0.5}
\newcommand{\blue}[1]{\textcolor{airforceblue}{#1}}
\title{\textit{Just Ask One More Time!}\\Self-Agreement Improves Reasoning of Language Models \\ in (Almost) All Scenarios}

\author{Lei Lin\textsuperscript{1\footnotemark[1]}~~~
Jiayi Fu\textsuperscript{1\footnotemark[1]}~~~
Pengli Liu \textsuperscript{1}~~~
Qingyang Li \textsuperscript{1}~~~
Yan Gong \textsuperscript{2\thanks{~~Equal Contribution.}}~~~
Junchen Wan\textsuperscript{1}~~~\\
\textbf{
Fuzheng Zhang\textsuperscript{1}~~~
Zhongyuan Wang\textsuperscript{1}~~~
Di Zhang\textsuperscript{1}~~~
Kun Gai\textsuperscript{1}~~~
}\\
\textsuperscript{1}Kuaishou Technology, Beijing, China\\
\textsuperscript{2}School of Computer Science and Engineering, Northeastern University, Shenyang, China\\
\texttt{\{linlei,fujiayi\}@kuaishou.com,\{wanjunchen,zhangfuzheng\}@kuaishou.com} \\}

\begin{document}
\maketitle
\begin{tikzpicture}[remember picture,overlay,shift={(current page.north west)}]
\node[anchor=north west,xshift=6.6cm,yshift=-2.0cm]{\scalebox{0.5}{\includegraphics[width=2cm]{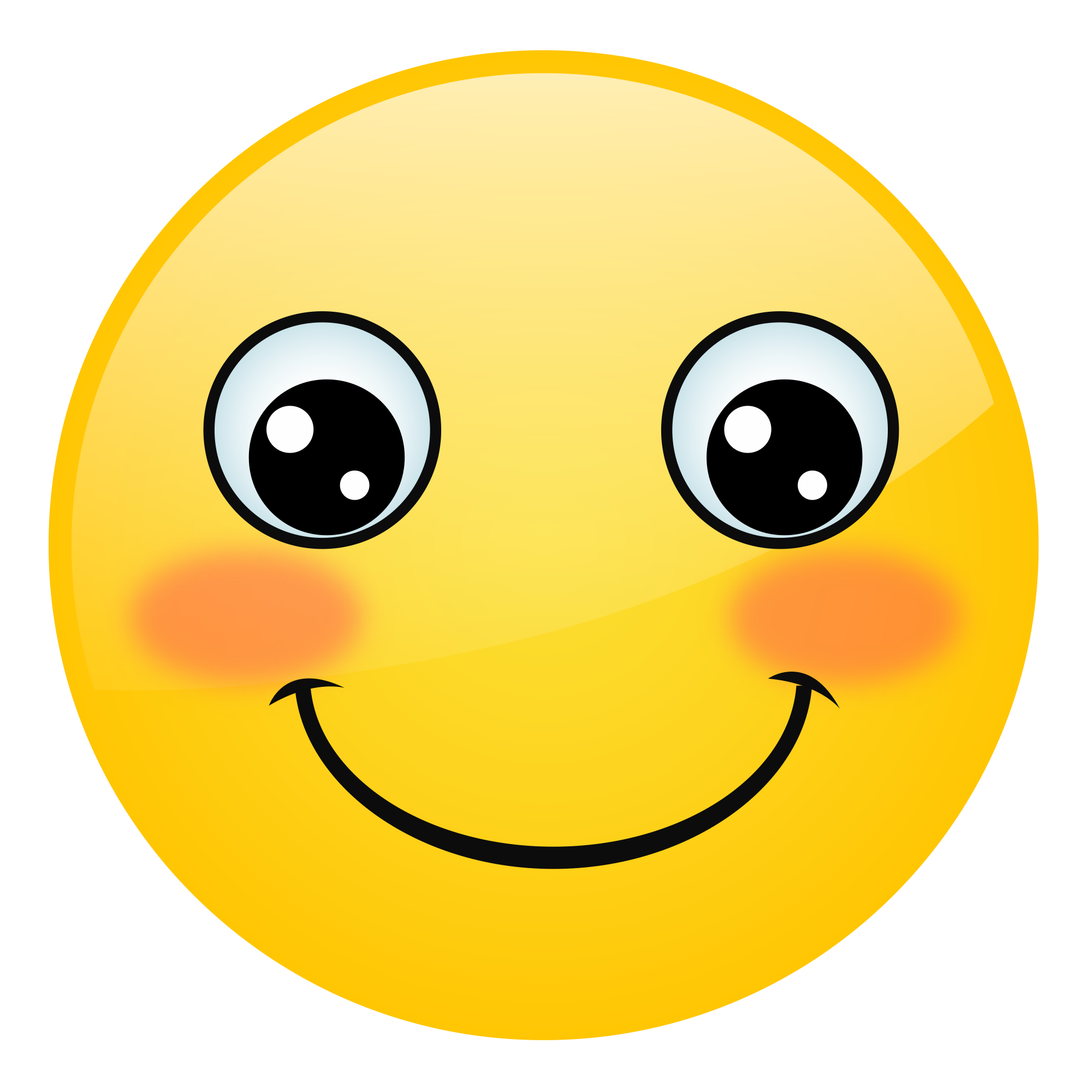}}};
\end{tikzpicture}
\begin{abstract}
\label{sec:abs}

Although chain-of-thought (CoT) prompting combined with language models has achieved encouraging results on complex reasoning tasks, the naive greedy decoding used in CoT prompting usually causes the repetitiveness and local optimality. To address this shortcoming, ensemble-optimization tries to obtain multiple reasoning paths to get the final answer assembly.
% , which avoids the repetitiveness and local optimality. 
However, current ensemble-optimization methods either simply employ rule-based post-processing such as \textit{self-consistency}, or train an additional model based on several task-related human annotations to select the best one among multiple reasoning paths, yet fail to generalize to realistic settings where the type of input questions is unknown or the answer format of reasoning paths is unknown. 
% Despite taking language models as multi-agents for debating can generalize to such cases, it is complex as the process of agents debating is iterative.
% , it may be difficult to equip with realistic language models products as the process of debating is iterative, resulting in a relatively delayed response for users. 
To avoid their limitations, we propose \textbf{Self-Agreement}, a generalizable ensemble-optimization method applying in almost all scenarios where the type of input questions and the answer format of reasoning paths may be known or unknown. Self-agreement firstly samples from language model's decoder to generate a \textit{diverse} set of reasoning paths, and subsequently prompts the language model \textbf{\textit{one more time}} to determine the optimal answer by selecting the most \textit{agreed} answer among the sampled reasoning paths. Self-agreement simultaneously achieves remarkable performance on six public reasoning benchmarks and superior generalization capabilities.

\end{abstract}

\section{Introduction}
\label{sec:intro}

% Reasoning, the process of making inference based on existing information or knowledge, is the core of human intelligence and essential for soling complex questions~\cite{yu2023nature}. As noted in OpenAI, ``In a casual conversation,
% the distinction between GPT-3.5 and GPT-4 can be subtle.The difference comes out when the complexity of the task
% reaches a sufficient threshold.''
Although large language models (LLMs) have revolutionized the natural language processing (NLP) landscape, their ability to solve challenging tasks (e.g., arithmetic, commonsense and symbolic reasoning) is often seen as a limitation, which is difficult to be overcome solely by scaling up the size of LLMs~\cite{rae2021scaling,srivastava2022beyond}. To address this issue,~\citet{wei2022chain} have proposed \textit{chain-of-thought} (CoT) prompting, which provides a few examples consisting of reasoning steps to prompt LLMs to generate intermediate reasoning steps towards final answers.
% that a person might employ in solving a complex task. 
It has been demonstrated that CoT prompting can elicit strong reasoning capabilities from LLMs, and achieve superior performance in solving complex tasks~\cite{wei2022chain}. However, the naive greedy decoding strategy used in CoT prompting usually causes the repetitiveness and local optimality.
% ~\citet{wang2022self}  propose self-consistency, to replace the naive greedy decoding used in chain-of-thought prompting to avoid the repetitiveness and local-optimality that plague greedy decoding.
% Therefore, designing prompts manually~\cite{wei2022chain,fu2022complexity,diao2023active} or automatically~\cite{kojima2022large,zhang2022automatic,shum2023automatic}, and example selection in chain-of-thoughts prompting~\cite{fu2022complexity,zhang2022automatic,shum2023automatic,diao2023active} have become a hot topic in NLP.
% A key technique they all use is the greedy decoding strategy, since reasoning tasks typically have fixed answers.
% scaling up LLM size alone has demonstrated to struggle on challenging tasks such as arithmetic, commonsense, and symbolic reasoning.

% Reasoning tasks typically have fixed answers, which is why most of previous work has generally considered greedy decoding approaches~\cite{radford2019language,wei2022chain}. On the contrary, 

This work studies \textit{ensemble-optimization}~\cite{qiao2022reasoning} in multi-step reasoning situations. Ensemble-optimization tries to obtain multiple reasoning paths to get the final answer assembly. It avoids the repetitiveness and local optimality that plague greedy decoding, while mitigating the stochasticity caused by a single sampled generation~\cite{wang2022self}.
% Such an approach is analogous to a fact that assessing whether or not a person really know how to solve a problem should depend on multiple generated reasoning paths by themself, since a person might acquire the wrong answer due to carelessness or other factors.
\begin{figure*}
    \centering
    \includegraphics[width=0.9\linewidth]{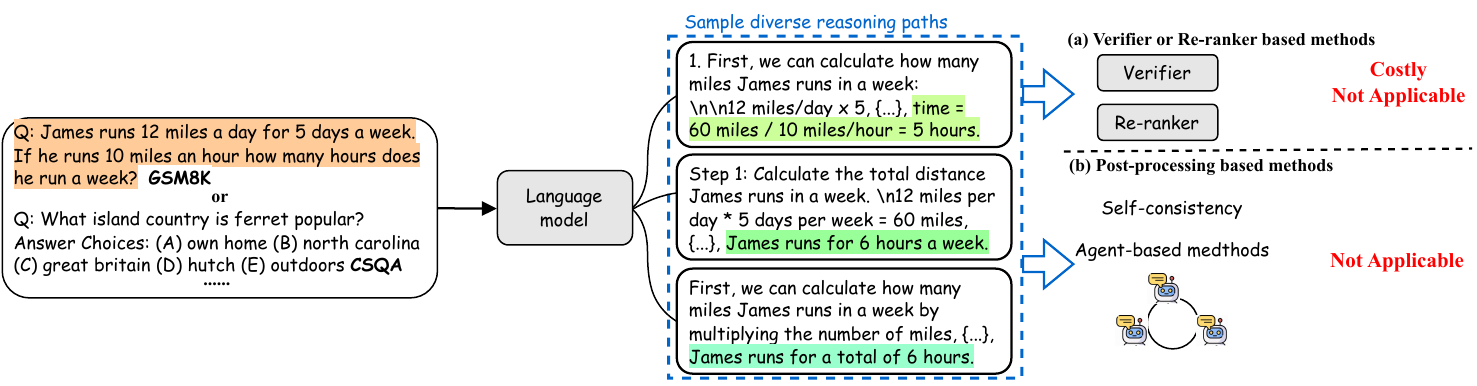}
    \caption{A simple example to illustrate the main limitations of existing ensemble-optimization methods. Both verifier or re-ranker based methods and post-processing based methods can only solve the question that belongs to a known task, yet fail to generalize to situations of the unknown type of questions.
    % Additionally, although agent based methods can generalize to such realistic situations, they are complex as the process of agents debating is iterative.
    }
    \label{fig:fig1}
\end{figure*}

Current ensemble-optimization methods predominantly fall into two categories, i.e., \textit{verifier or re-ranker based methods} and  \textit{post-processing based methods}.
% and \textit{agent based methods}, respectively. 
Verifier or re-ranker based methods either train an additional verifier~\cite{cobbe2021training,li2023making} or train a re-ranker based on task-related human annotations~\cite{thoppilan2022lamda} to select the best generation among multiple generations. Post-processing based methods take a majority vote among all generated reasoning paths~\cite{wang2022self,du2023improving,liang2023encouraging,chen2023reconcile} or \textit{top K} complex reasoning paths~\cite{fu2022complexity}, and then choose the optimal answer that receives the most votes.
% Agent based methods take language models as multi-agents to simulate the process of debating over multiple rounds to arrive at the common final answer~\cite{du2023improving,liang2023encouraging,chen2023reconcile}, where one agent's input usually depends on other agents' outputs.

While the success of ensemble-optimization work, we show that there are two major limitations. Figure~\ref{fig:fig1} illustrates the main limitations of existing ensemble-optimization methods. On one hand, although verifier or re-ranker based methods have favorable generalization ability to some extent as they impose no restrictions on answer formats of reasoning paths, but they can only be applied in scenario where the type of questions is already known, yet fail to generalize to scenarios where the type of questions is unknown. Besides, it is costly to train such an additional model. On the other hand, post-processing based methods can only be applied in scenario where both the type of questions and answer formats of reasoning paths are already known.

% agent based methods have favorable generalization ability as they impose no restrictions on types of questions and answer formats of reasoning paths, but they are complex due to multi-round interactions taking language models as multi-agents.
% to arrive the common final answer. 
% On the other hand, both verifier or re-ranker based methods and post-processing based methods can only be applied in scenario where the type of questions is already known, yet fail to generalize to scenarios where the type of questions is unknown. Besides, it is costly to train such an additional model.
% to select the best one among multiple reasoning paths.

\begin{figure*}
    \centering
    \includegraphics[width=0.95\linewidth]{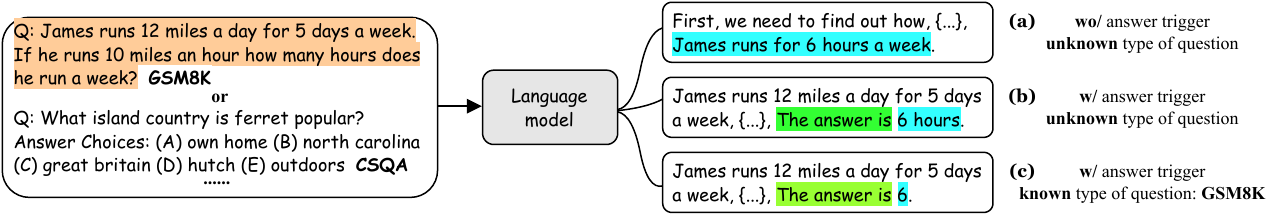}
    \caption{(a), (b) and (c) represent three major different cases covering almost all possible situations. The first scenario (a) is that both the type of question and answer format are unknown, and the second one (b) is that the type of question is unknown while the answer format is known. The third one (c) is that the type of question and answer format are already known (i.e., the type of questions belongs to GSM8K~\cite{cobbe2021training}, and the answer has the trigger (e.g., \textit{The answer is})).}
    \label{fig:fig2}
\end{figure*}

Nevertheless, in practical applications, language models often encounter situations of unknown type of questions or different answer formats, where it cannot be clearly identified which task the questions belong to and what answer format of the reasoning paths appears to be. Figure~\ref{fig:fig2} shows a simple example to illustrate three major cases covering almost all possible situations. Briefly, we categorize the types of questions into known and unknown, i.e., whether we can identify in advance that the question belongs to a particular task, and answer formats into with (w/) and without (wo/) trigger (e.g., \textit{The answer is}).\footnote{It is natural to simulate if the type of questions is known by whether or not to mix different reasoning tasks. Besides, we simulate whether the reasoning path has a trigger via using few-shot CoT or zero-shot CoT. We employ zero-shot CoT without 2nd answer extraction phase to simulate the reasoning paths with no trigger.} So, the first case (a) is that both the type of question and answer format are unknown, the second one (b) is that the type of question is unknown while the answer format is known, and the third one (c) is that the type of question and answer format are already known. When encountering situations of unknown type of questions and answer formats (i.e., the first or second case), it is neither reasonable to train an additional model given several task-related human annotations nor possible to manually identify which task it refers to,
% and determine the reasoning paths have a fixed answer trigger in advance,
% and then employ post-processing based methods, 
not to mention that the question encountered in real-world scenarios is not even from a pre-defined set of tasks.
Besides, the answer formats of reasoning paths in real-world are ever-changing. However, existing ensemble-optimization studies commonly assume that both the type of questions and answer format of reasoning paths are already known, which is contradictory to real-world scenarios and has limited application values.
% Besides, taking language models as multi-agents to perform debating
% may be difficult to equip with realistic language models products, such as KwaiYiiMath~\cite{fu2023kwaiyiimath}, Baichuan 2~\cite{yang2023baichuan}, etc., since it 
% is complex as the process of agents debating is conducted iteratively.
% , resulting in a relatively delayed response for users. 
Therefore, one natural question can be raised: how to design a \textit{simple} ensemble-optimization method
% equipped with realistic language models products 
to solve reasoning tasks in \textit{almost all situations}?

To this end, we propose \textit{self-agreement}, a new ensemble-optimization scheme that
% can equip with realistic language models products to 
further improves reasoning performance of language models in almost all possible scenarios by a large margin. 
% Self-agreement is designed to select the most agreed answer based on a diverse set of sampled reasoning paths even in unknown type of questions by itself. 
We get insights from the majority vote design used in~\citet{wang2022self}, and start from the intuition that assessing whether or not a person really knows how to solve a question should depend on multiple reasoning paths generated by themself, since a person might acquire the wrong answer due to carelessness or other factors.

\begin{figure*}
    \centering
    \includegraphics[width=0.95\linewidth]{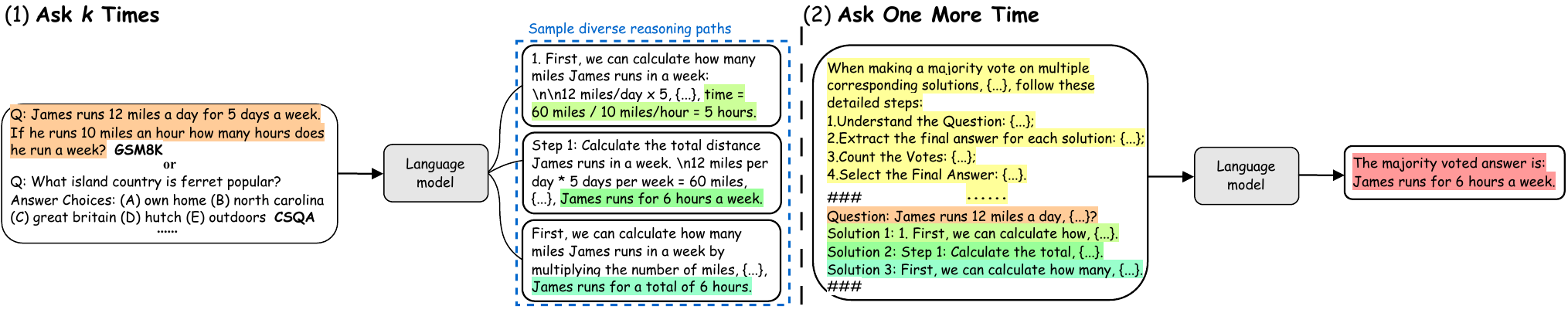}
    \caption{The self-agreement method contains two stages: (1) \textit{ask k times}: sample from the language model itself \textit{k} times to generate $k$ diverse reasoning paths using few-shot CoT~\cite{wei2022chain} or zero-shot CoT~\cite{kojima2022large}; and (2) \textit{ask one more time}:  select the most agreed answer based on 
    \textit{k} reasoning paths by language model itself. Yellow denotes the carefully designed prompt proposed in this work (see Table~\ref{table:t5} in the Appendix), and red denotes the most agreed answer generated by the language model itself. 
    % The carefully designed prompt is given in Table~\ref{table:t5} in the Appendix.
    }
    \label{fig:fig3}
\end{figure*}

% The fixed reasoning question type is that the answer to the reasoning question has a answer pattern (e.g., \textit{the answer is}) and known answer types (e.g., the final answer to the reasoning question is arabic numerals). The semi-fixed reasoning question type is that the answer to the reasoning question has a answer pattern and unknown answer types (e.g., the final answer to the reasoning question may be arabic numerals, or options, etc). The unfixed reasoning question type is that the answer to the reasoning question has unknown answer types without a answer pattern.
Figure~\ref{fig:fig3} illustrates the self-agreement method with an example. Self-agreement comprises two phases: Firstly, we prompt the language model with \textit{few-shot CoT}~\cite{wei2022chain} (w/ answer trigger) or \textit{zero-shot CoT}~\cite{kojima2022large} (wo/ answer trigger). Then, we sample from language model's decoder to generate a \textit{diverse} set of reasoning paths. We name this phase \textit{ask k times}, where \textit{k} is the size of the diverse set; Secondly, we prompt the language model \textit{one more time} to determine the optimal answer by selecting the most agreed answer among the sampled reasoning paths. We name this phase \textit{ask one more time}. Such an approach is analogous to the fact that the most agreed answer selected from multiple reasoning paths is likely to be the correct answer, since most reasoning paths generated by themself have a high probability to arrive at the same correct answer if a person actually knows how to solve it. Therefore, we refer to our approach as \textbf{Self-Agreement}, which stands for the fact. Self-agreement is simple yet effective, completely unsupervised, requires no additional human annotations or auxiliary models, and avoids any additional training or fine-tuning. \textit{To the best of our knowledge, we are the first to propose the real-world scenarios in ensemble-optimization studies, and design a simple ensemble-optimization method applying in almost all scenarios.}\footnote{After completion of this work, we find that~\citet{chen2023universal} share almost the same idea with us, which is archived after our work. We provide discussion with it in Appendix~\ref{sec:dis}.}

To validate the efficacy of self-agreement, we conduct experiments on six public reasoning benchmarks covering arithmetic, commonsense and symbolic reasoning. Experimental results show that self-agreement improves reasoning performance of language models by a striking margin across all tasks in all three scenarios. Moreover, our method also exhibits versatility across various language models and model sizes, task settings, reasoning paths with different diversity, and prompting methods, highlighting its broad applicability.
% In addition, self-agreement also attains superior generality.
% in mixed-task scenarios.
% simultaneously attain remarkable performances and superior generality. 
% Notably, self-agreement achieves better performance on GSM8K~\cite{cobbe2021training} (+12.9\% absolute accuracy gains) over self-consistency~\cite{wang2022self}.

% We first prompt the language model with \textit{any one of the prompting techniques} to generate a \textit{diverse} set of reasoning paths using sample strategies, such as zero-shot CoT, few-shot CoT, etc. We refer to this stage as \textit{ask $k$ times}, where $k$ is the size of the diverse set. Then, we prompt the language model one more time to select the most agreed answer based on $k$ reasoning paths, given a carefully designed prompt.\footnote{The prompt for reasonig tasks is given in Table~\ref{table:t4}.} We refer this stage as \textit{ask one more time}. Self-agreement is far simpler than prior verifier or re-ranker based methods and agent based methods. Instead, self-agreement is entirely unsupervised, works off-the-shelf with pre-trained language models, requires no additional human annotation, and avoids any additional training, auxiliary models or fine-tuning.

% we \textit{ask the language model one more time} to select the most agreed answer by designing prompt, since each reasoning path might lead to a different final answer.

% 优点，讲相关工作没有这个优点

% 实验，小图表明当前这个方向的工作的缺点

\section{Related Work}
\label{sec:rw}

% \paragraph{Reasoning in Language Models and Prompting.}
% \paragraph{In-context Learning.}
\noindent\textbf{In-context Learning}
Language models have revolutionized a wide range of NLP tasks, where scaling up the model size is one of the key ingredients~\cite{vaswani2017attention,kenton2019bert,raffel2020exploring,brown2020language,rae2021scaling,chowdhery2022palm,thoppilan2022lamda}. The success of LLMs is often attributed to \textit{emergent abilities} when the model reaches a sufficient scale~\cite{wei2022emergent}. That is, the model can follow the format of given prompts (typically a few task-specific examples) thus solving the corresponding tasks (also referred as \textit{in-context learning}). The method of conditioning LLMs is called ``prompting''~\cite{liu2023pre}, which can be categorized into two main directions: \textit{few-shot prompting} (a few examples as the prompt) and \textit{zero-shot prompting} (instructions describing the task as the prompt). Prompting allows a single model to carry out various tasks universally. Due to its superior benefits, there are studies~\cite{liu2021makes,lu2021fantastically,lu2022dynamic} further investigate how to improve the performance of in-context learning. Specifically, different wording or order of given examples may lead to performance fluctuations~\cite{zhao2021calibrate,webson2021prompt}. This work takes an important step forward in multi-step reasoning by showing the critical role of language model itself in the process of selecting the most agreed answer based on multiple reasoning paths.

\noindent\textbf{Reasoning with Language Models}
% \paragraph{Reasoning in Language Models.}
Reasoning, the process of making inference based on existing information or knowledge, is the core of human intelligence and essential for solving complex questions~\cite{yu2023nature}. In contrast to the excellent performance of LLMs in simple and single-step tasks,
% with few-shot prompting or zero-shot prompting, 
language models (even 100B or more parameters) are demonstrated to struggle at solving challenging tasks required multi-step reasoning~\cite{rae2021scaling,srivastava2022beyond}.
% , such as arithmetic, commonsense and symbolic reasoning~\cite{rae2021scaling,srivastava2022beyond}. 
To address this issue,~\citet{nye2021show} have proposed to decompose multi-step reasoning problems into intermediate steps before obtaining final answers. Furthermore,~\citet{wei2022chain} have proposed chain-of-thought prompting, which elicits this reasoning process from language models.
% and achieves superior performance. 
Since then, designing prompts manually~\cite{wei2022chain,fu2022complexity,diao2023active} or automatically~\cite{kojima2022large,zhang2022automatic,shum2023automatic}, example selection in CoT prompting~\cite{fu2022complexity,zhang2022automatic,shum2023automatic,diao2023active}, and data augmentation~\cite{fu2023kwaiyiimath,yue2023mammoth,yu2023metamath} have become a hot topic in NLP. Our work sits in the context of zero-shot CoT and few-shot CoT, and proposes a new ensemble-optimization method that substantially outperforms the original zero-shot CoT and few-shot CoT, respectively.
% Compared to prior work, self-agreement is applicable to a wide range of reasoning tasks without any additional supervision or fine-tuning, while still substantially improving the performance of the few-shot CoT~\cite{wei2022chain} and zero-shot CoT~\cite{kojima2022large}.
% the CoT prompting approach proposed in~\citet{wei2022chain}.

% \paragraph{Ensemble-Optimization in Language Models.}
\noindent\textbf{Ensemble-Optimization in Reasoning with Language Models} Ensemble-optimization tries to obtain multiple reasoning paths to get the final answer assembly~\cite{qiao2022reasoning}. 
% Current ensemble-optimization methods predominantly fall into three categories, i.e., \textit{verifier or re-ranker based methods}, \textit{post-processing based methods} and \textit{agent based methods}, respectively. 
We view most ensemble-optimization work as different implementations of \textit{the majority voted idea}.\footnote{Details refer to corresponding papers.} For example, self-consistency~\cite{wang2022self} takes the majority vote over sampled reasoning paths, while multi-agents (debate)~\cite{du2023improving} obtains the majority voted answer over multiple agents' outputs. It is clear that the essence of them is based on the majority voted design, and our work is in lines with it, \textit{but we explore how to allow language model itself to achieve the overall procedure}. So, the main difference between them lies in two major steps consisting of the process of the majority vote, i.e., answer extraction and answer comparison. To be specific, self-consistency and multi-agents (debate) are rule-based and not generalizable method to extract and compare answers.
% , while multi-agents (debate) has a generalizable answer extraction yet rule-based answer comparison way. 
However, self-agreement is an entirely generalizable method, whether extracting answers or comparing answers. 

\section{Methodology}
\label{sec:met}

Existing ensemble-optimization studies~\cite{wang2022self,fu2022complexity} commonly assume that both the type of questions fed to the model and answer format of reasoning paths outputted by the model are already known, and conduct evaluations on the questions from the same dataset. 
% Then, they conduct evaluations on the questions from the same task or dataset. 
However, a more realistic setting is that the type of input questions or answer format of reasoning paths is unknown and they come in an arbitrary manner. To address such scenarios, one natural idea is that can we allow \textit{the language model itself} to select the best generation among multiple generations, since it has favorable generalization ability, i.e., no restrictions on the types of questions and answer formats of reasoning paths, and avoids any additional training, auxiliary models or fine-tuning. Beforehand, we need to figure out \textit{what language models are good at for}? 

\begin{table*}[t]
  \centering
  \small
  \resizebox{1.0\linewidth}{!}{
  \begin{tabular}{@{}ll  @{\hspace{2pt}} cccccccc}
      \toprule
      \bf Model & & \multicolumn{3}{c}{\textit{Arithmetic}} & \multicolumn{2}{c}{\textit{Commonsense}} & \multicolumn{1}{c}{\textit{Symbolic}} & \multirow{2}{*}{\bf Avg} \\
      \cmidrule(r){3-5}
      \cmidrule(r){6-7}
      \cmidrule(r){8-8}
       & & \bf GSM8K & \bf MultiArith & \bf SVAMP & \bf CSQA & \bf ARC-c  & \bf Letter \\\midrule
       \multicolumn{10}{c}{\textit{Both the type of questions and the answer format are unknown (\textbf{the first scenario})}} \\
       \midrule
       \multirow{3}{*}{GPT-3.5-turbo} & Zero-Shot CoT & 48.5 & 93.0 & 73.5 & 71.0 & 81.5 & 81.0 & 74.8 \\ 
                  % & Zero-shot CoT$^{\dagger}$ & 45.7 & 93 & 74 & 49 & 57.5 & 68.5 & 83 & 67.2 \\
                  & Self-Consistency & N/A &  N/A & N/A & N/A & N/A & N/A & N/A \\
                  & Self-Agreement & \bf 58.5 (\blue{+10.0}) &  \bf 93.4 (\blue{+0.4}) & \bf 79.6 (\blue{+6.1}) & \bf 77.8 (\blue{+6.8}) & \bf 87.7 (\blue{+6.2}) & \bf 83.8 (\blue{+2.8}) & \bf 80.1 (\blue{+5.3}) \\
                  % & Self-Agreement$^{\dagger}$ & 72.5 & 99 & 83.5 & 78.5 & 88.5 & 81 & 65 & 81.1 \\
                  
        \midrule
    \multirow{3}{*}{Llama-2-13B-Chat} 
    % & Zero-Shot CoT & 25.0 & 78.5 & 50.5 & 58.0 & 66.0 & 31.5 & 51.6 \\ 
    & Zero-Shot CoT & 26.0 & 77.5 & 50.5 & 57.5 & 66.5 & 31.0 & 51.5 \\
                   % & Zero-Shot CoT$^{\dagger}$ & 27.0 & 72.5 & 46.0 & 57.0 & 57.5 & 46.0 & 61.5 & 52.5 \\
                  & Self-Consistency & N/A &  N/A & N/A & N/A & N/A & N/A & N/A \\
                   % & Self-Agreement-256$^{\dagger}$ & 36.6 & 74.0 & 57.0 & 64.0 & 62.0 & 31.5 & 48.0 & 53.3 \\
                  % & Self-Agreement-256 & \bf 12.7 & \bf 28.4 & \bf 25.0 & \bf 58.4 & \bf 59.1 & \bf 3.0 & \bf 13.5 & \bf 28.6 \\
                  % & Self-Agreement-512 & \bf 14.4 & \bf 42.5 & \bf 38.0 & \bf 61.6 & \bf 60.7 & \bf 15.2 & \bf 13.4 & \bf 35.1 \\
                  & Self-Agreement & \bf 31.7 (\blue{+5.7}) & \bf 77.8 (\blue{+0.3}) & \bf 56.3 (\blue{+5.8}) & \bf 65.0 (\blue{+7.5}) & \bf 68.5 (\blue{+2.0}) & \bf 44.5 (\blue{+13.5}) & \bf 57.3 (\blue{+5.8}) \\
        \midrule
        \multicolumn{10}{c}{\textit{The type of questions is unknown and the answer format is known (\textbf{the second scenario})}} \\
        \midrule
      \multirow{4}{*}{GPT-3.5-turbo} & Mixed-Few-Shot CoT & 69.0 & 97.5 & 76.5 & 71.0 & 86.0 & 83.0 & 80.5 \\ 
                  & Self-Consistency & 68.0 &  98.1 & 76.6 & 71.2 & 83.1 & 79.1 & 79.3 \\
                  & Multi-Agents (Debate) & 80.2 & 96.0 & 76.2 & 69.0 & 78.4 & 61.3 & 76.9 \\
                  & Self-Agreement & \bf 81.3 (\blue{+12.3}) & \bf 99.1 (\blue{+2.6}) & \bf 83.7 (\blue{+7.2}) & \bf 75.4 (\blue{+4.4}) & \bf 87.8 (\blue{+1.8}) & \bf 88.9 (\blue{+6.9}) & \bf 86.0 (\blue{+5.5}) \\
                  \midrule
        \multirow{4}{*}{Llama-2-13B-Chat} & Mixed-Few-Shot CoT & 31.5 & 84.0 & 61.0 & 66.5 & 68.0 & 19.0 & 55.0 \\ 
                  & Self-Consistency & 32.8 & 83.9  & 56.7 & 68.0 & 68.1 & 16.0 & 54.5 & \\ 
                  % & Self-Agreement-256 & \bf 34.7 & \bf 83.1 & \bf 56.9 & \bf 70.2 & \bf 67.1 & \bf 0.8 & \bf 59.0 & \bf 53.1 \\
                  % & Self-Agreement-512 & \bf 38.4 & \bf 91.3 & \bf 63.2 & \bf 70.7 & \bf 69.0 & \bf 17.3 & \bf 63.0 & \bf 59.0 \\
                  & Multi-Agents (Debate) & \bf 44.0 & 79.0 & 54.0 & 52.0 & 56.5 & \bf 27.0 & 52.1\\
                  & Self-Agreement & 41.5 (\blue{+10.0}) & \bf 94.7 (\blue{+10.7}) & \bf 66.8 (\blue{+5.8}) & \bf 70.2 (\blue{+3.7}) & \bf 70.6 (\blue{+2.6}) & 23.1 (\blue{+4.1}) & \bf 61.2 (\blue{+6.2}) \\
      \bottomrule
  \end{tabular}
  }
  \caption{
  Self-agreement, when applied on GPT-3.5-turbo and Llama-2-13B-Chat. Our performance gain (\blue{\textbf{+blue}}) is computed over the mixed-few-shot CoT or zero-shot CoT~\cite{kojima2022large}, which is our primary baseline. Our method substantially increases the performance over mixed-few-shot CoT and zero-shot CoT. 
  The best performance across each model for each task is shown in bold.
  % , with an average \textbf{\blue{+5.3}} and \textbf{\blue{+5.3}} gains on GPT-3.5-turbo respectively.
  }
  \label{table:t1}
\end{table*}
Previous work has demonstrated that language models are good answer extractors given specific prompts~\cite{kojima2022large,zhou2022large,yang2023large}, and also suitable for comparing the consistency of final answers extracted from multiple reasoning paths given specific prompts~\cite{yang2023harnessing,yang2023skymath}, as the extracted final answers in reasoning tasks tend to be short and semantically straightforward sentences or phrases. For instance, as shown in Figure~\ref{fig:fig3}, for language models, \textit{James runs for 6 hours a week} and \textit{James runs for a total of 6 hours} extracted from solution 2, 3 respectively represent \textbf{the same final answer}. 
% Thus, language models can extract the final answer of each reasoning path, and compare the consistency among multiple final answers. 
Such behaviors are often attributed to \textit{emergent abilities}~\cite{wei2022emergent}. 
That is, the model can understand and follow the format of given prompts thus solving the corresponding tasks.

Motivated by the above findings and the majority vote design used in~\citet{wang2022self}, we propose the following \textit{self-agreement} method that allows language model itself to extract final answers of multiple reasoning paths, count the votes of different answers, and then select the majority voted answer. Specifically, firstly, we prompt the language model with \textit{few-shot CoT}~\cite{wei2022chain} (w/ answer trigger) or \textit{zero-shot CoT}~\cite{kojima2022large} (wo/ answer trigger). Then, we sample from language model's decoder to generate a \textit{diverse} set of reasoning paths. We refer to this stage as \textit{ask k times}, where \textit{k} is is the size of the diverse set. Self-agreement is compatible with reasoning paths to the questions with different diversity.\footnote{For example, in temperature sampling strategies~\cite{ficler2017controlling}, higher values like 0.8 will make the output more random, while lower values like 0.2 will make it more focused and deterministic.}
% most existing sampling algorithms, including temperature sampling~\cite{ficler2017controlling}, and nucleus sampling~\cite{holtzman2019curious}, etc. 
Secondly, we prompt the language model \textit{one more time} to determine the optimal answer by selecting the most agreed answer among the sampled reasoning paths. We refer to this stage as \textit{ask one more time}. Figure~\ref{fig:fig3} shows the overall procedure of our approach.

% \noindent\textbf{Ask \textit{k} Times.} 
% \paragraph{Ask \textit{k} Times}
\subsection{Ask \textit{k} Times}
\label{sec:me_sub1}

% As shown in Figure~\ref{fig:fig2}, there are three major cases covering almost all possible situations while existing ensemble-optimization studies commonly conduct experiments on the third case. 
We simulate whether the type of questions is known or not by whether or not we mix tasks,\footnote{Nearing completion of this work, we find that~\citet{zou2023meta} have applied CoT prompting to mixed-task scenarios to simulate the real-world applications, which share the same idea with us while we propose it in ensemble-optimization studies.} and whether the answer format of reasoning paths is known or not by utilizing few-shot CoT~\cite{wei2022chain}
% \footnote{Note that we use the same prompts used in~\citet{wei2022chain} for the first case, and the general prompt by mixing examples used in~\citet{wei2022chain} for the second case.}
or zero-shot CoT~\cite{kojima2022large}. In this stage, we first modify the input question $\mathbf{x}$ into a \textit{prompt} $\mathbf{x}^{\prime}$. The prompt $\mathbf{x}^{\prime}$ would be ``Q: \texttt{[X]}. A: '' with a few hand-crafted examples or ``Q: \texttt{[X]}. A: Let's think step by step.'' if we employ few-shot CoT or zero-shot CoT, where \texttt{[X]} is an input slot for $\mathbf{x}$. The same prompted text $\mathbf{x}^{\prime}$ is then fed into language models and generate multiple sentences $\mathbf{z}_{i}$ via sampling strategies, where $i=1, \ldots, m$ indexes the $m$ candidate outputs.

\begin{table*}[t]
  \centering
  \small
  \resizebox{0.95\linewidth}{!}{
  \begin{tabular}{@{}ll  @{\hspace{2pt}} cccccc}
      \toprule
      \bf Model & & \multicolumn{3}{c}{\textit{Arithmetic}} & \multicolumn{2}{c}{\textit{Commonsense}} & \multicolumn{1}{c}{\textit{Symbolic}} \\
      \cmidrule(r){3-5}
      \cmidrule(r){6-7}
      \cmidrule(r){8-8}
       & & \bf GSM8K & \bf MultiArith & \bf SVAMP & \bf CSQA & \bf ARC-c  & \bf Letter \\\midrule
       \multicolumn{8}{c}{\textit{Both the type of questions and the answer format are known (\textbf{the third scenario})}} \\
        \midrule
      \multicolumn{2}{@{}l}{Previous finetuning SOTA } & 55.0$^a$ & 60.5$^b$ & 57.4$^c$ & \bf 91.2$^d$ & 75.0$^e$ & N/A \\\midrule
      \multirow{2}{*}{LaMDA$^\dagger$} & Few-Shot CoT & 17.1 & 51.8 & 38.9 & 57.9 & 55.1 & 8.2 \\
      & Self-Consistency & 27.7 & 75.7 & 53.3 & 63.1 & 59.8 & 8.2 \\
      \midrule
      \multirow{2}{*}{PaLM$^\dagger$} & Few-Shot CoT & 56.5 & 94.7 & 79.0 & 79.0 & 85.2 & 65.8 \\
      & Self-Consistency & 74.4 & \bf 99.3 & \bf 86.6 & 80.7 & \bf 88.7 & 70.8 \\
      \midrule
      \multirow{2}{*}{Minerva$^\dagger$} & Few-Shot CoT & 58.8 & - & - & - & - & - \\
      & Self-Consistency & 78.5 & - & - & - & - & - \\
      % \multicolumn{2}{@{}l}{Minerva$^\dagger$}  & 58.8 & - & -\\
     \midrule
      \multirow{4}{*}{GPT-3.5-turbo} & Few-Shot CoT & 70.0 & 98.2 & 82.0 & 74.5 & 85.4 & 80.6 \\ 
                  % & Self-Consistency & 81.9 & 99.2 & 86.5 & 79.4 & 87.1 & \bf 82.4 \\
                  & Self-Consistency & 80.3 & 99.2 & 85.9 & 79.0 & 87.0 & \bf 81.9 \\ 
                  & USC & 76.8 & 98.2 & 83.5 & 48.9 & 73.9 & 79.4 \\ 
                  & Self-Agreement & \bf 82.4 (\blue{+12.4}) & 99.0 (\blue{+0.8}) & 86.0 (\blue{+4.0}) & 79.4 (\blue{+4.9}) & 86.8 (\blue{+1.4}) & 81.0 (\blue{+0.4}) \\ 
                  
                  % & Self-Agreement-R & \bf 83.1 (\blue{+13.1}) & 99.0 (\blue{+0.8}) & 85.8 (\blue{+3.8}) & \bf 80.1 (\blue{+5.6}) & 86.6 (\blue{+1.2}) & 81.7 (\blue{+1.1}) & 99.7 (\blue{+4.7}) \\ 
    % \midrule
    %   \multirow{3}{*}{Llama-2-70B-Chat} & Few-Shot CoT & 48.1 & - & - & 77.1 & 85.4 & 80.6 \\ 
    %               & Self-Consistency & 59.7 & - & - & 78.1 & 87.1 & \bf 82.4 \\ 
    %               & Self-Agreement & \bf 61.0 (\blue{+12.4}) & - & - & 77.7 (\blue{+4.9}) & 86.8 (\blue{+1.4}) & 81.0 (\blue{+0.4}) \\ 
    % \midrule
    %   \multirow{3}{*}{GPT-4} & Few-Shot CoT & 90.8 & - & - & 86.9 & 85.4 & 80.6 \\ 
    %               & Self-Consistency & 93.3 & - & - & 88.1 & 87.1 & \bf 82.4 \\ 
    %               & Self-Agreement & \bf 93.9 (\blue{+12.4}) & - & - & 87.9 (\blue{+4.9}) & 86.8 (\blue{+1.4}) & 81.0 (\blue{+0.4}) \\ 
                  % & Self-Agreement-R & \bf 83.1 (\blue{+13.1}) & 99.0 (\blue{+0.8}) & 85.8 (\blue{+3.8}) & \bf 80.1 (\blue{+5.6}) & 86.6 (\blue{+1.2}) & 81.7 (\blue{+1.1}) & 99.7 (\blue{+4.7}) \\ 
      \bottomrule
  \end{tabular}
  }
  \caption{
  Self-agreement, when applied on GPT-3.5-turbo.
  $\dagger$ models are not publicly accessible, and the numbers are obtained from their papers. 
  Our performance gain (\blue{\textbf{+blue}}) is computed over few-shot CoT~\cite{wei2022chain}, which is our primary baseline.
  Our methods substantially increase the performance over~\citet{wei2022chain}, with an average \textbf{\blue{+4.0}} gain on GPT-3.5-turbo. 
  The previous finetuing SOTA baselines are obtained from:
  $a$: GPT-3 175B finetuned plus an additional 175B verifier~\cite{cobbe2021training},
  $b$: Relevance and LCA operation classifier~\cite{roy-roth-2015-solving},
  $c$:~\cite{pi2022reasoning},
  $d$: DeBERTaV3-large + KEAR~\cite{xu2021human},
  $e$: UnifiedQA-FT~\cite{khashabi2020unifiedqa}. The best performance for each task is shown in bold.
  }
  \label{table:t2}
\end{table*}
% \paragraph{Ask One More Time}
\subsection{Ask One More Time}
\label{sec:me_sub2}

After sampling multiple reasoning paths $\mathbf{z}_{i}$ from the model's decoder, self-agreement first extracts the final answers $\mathbf{a}_{i}$ of each reasoning path, and then selects the most ``agreed'' answer among the final answer set by taking a majority vote over $\mathbf{a}_{i}$.
% , i.e., $\argmax_a\sum\nolimits_{i=1}^m \mathbbm{1} (\mathbf{a}_i = a)$.
The overall process can be achieved by prompting language model itself one more time. The carefully designed prompt is given in Table~\ref{table:t5} in the Appendix respectively. Specifically, we first modify the input question $\mathbf{x}$ and its multiple reasoning paths $\mathbf{z}_{i}$ into a \textit{prompt} $\mathbf{s}$. Prompted text $\mathbf{s}$ is then fed into language models and generate subsequent sentence $\mathbf{v}$. We can use any decoding strategy, but we use greedy decoding in this stage for the simplicity.

% We first prompt the language model with \textit{few-shot CoT}~\cite{wei2022chain} (w/ answer trigger) or \textit{zero-shot CoT}~\cite{kojima2022large} (wo/ answer trigger) to generate a \textit{diverse} set of reasoning paths using sample strategies. We refer to this stage as \textit{ask $k$ times}, where $k$ is the size of the diverse set. Then, we prompt the language model one more time to select the most agreed answer based on $k$ reasoning paths, given a carefully designed prompt. We refer this stage as \textit{ask one more time}. For instance, as shown in Figure~\ref{fig:fig2}, the model might produce an incorrect reasoning path or make a mistake in one of the reasoning steps (Solution 3) like humans make careless mistakes, but it actually knows how to solve this math question because it obtains two correct answers of three reasoning paths. That is, we hypothesize that assessing whether or not a language model actually knows how to solve a problem should depend on multiple generated reasoning paths by themself.

\section{Experiments}
\label{sec:exp}

We carry out a series of experiments to confirm the efficacy of our method on three scenarios.
% mixed-task or single-task scenarios. 
Our findings indicate that across a wide range of tasks, scenarios, models, and prompting methods, self-agreement generally enhances the reasoning performance of language models.
% We find that self-agreement simultaneously achieves remarkable performances on the third scenario and superior generalization capabilities on the first and second scenarios. 
We introduce experimental setup in~\cref{sec:expset}, main results in~\cref{sec:mar}, and analysis in~\cref{sec:analysis}. See Appendix~\ref{sec:app_ed} and~\ref{sec:aexp} for more experimental details and additional experiments.

\subsection{Experimental Setup}
\label{sec:expset}

\noindent\textbf{Tasks and Datasets} We evaluate self-agreement on six public reasoning benchmarks for a fair comparison with existing methods.\footnote{We use the test split for all tasks if the labels are available for evaluation. For CommonsenseQA, we use the dev split.} These benchmarks can be divided into three categories of reasoning tasks: (i) \textbf{Arithmetic Reasoning} For these tasks, we use GSM8K~\cite{cobbe2021training}, SVAMP~\cite{patel2021nlp}, and MultiArith~\cite{roy2016solving}; (ii) \textbf{Commonsense Reasoning} We evaluate two commonsense reasoning tasks: CommonseQA (CSQA)~\cite{talmor2018commonsenseqa} and the AI2 Reasoning Challenge (ARC)~\cite{clark2018think}. The ARC dataset is divided into two sets: a challenge set (denoted as ARC-c), 
% containing questions that both retrieval-based and word co-occurrence algorithms answered incorrectly, 
and an easy set (denoted as ARC-e). We evaluate the effectiveness of our method on ARC-c; (iii) \textbf{Symbolic Reasoning} We choose last letter concatenation (e.g., the input is ``Elon Musk'' and the output should be ``nk'') from~\citet{wei2022chain}. To simulate the first and second scenarios, we select 200 examples from each reasoning task randomly and then mix them, since the cost of \textit{ask k times} stage is heavily expensive.
% Note that we sample 200 examples from each reasoning benchmark randomly to evaluate the effectiveness of our method in mixed-task scenario.
% \begin{itemize}
%     \item \textbf{Arithmetic Reasoning} For these tasks, we use GSM8K~\cite{cobbe2021training}, SVAMP~\cite{patel2021nlp}, and MultiArith~\cite{roy2016solving}.
%     \item \textbf{Commonsense Reasoning} We choose CommonseQA~\cite{talmor2018commonsenseqa} and the AI2 Reasoning Challenge (ARC)~\cite{clark2018think}. The ARC dataset is divided into two sets: a challenge set (denoted as ARC-c), containing questions that both retrieval-based and word co-occurrence algorithms answered incorrectly, and an easy set (denoted as ARC-e). We evaluate the effectiveness of our method on ARC-c.
%     \item \textbf{Symbolic Reasoning} We evaluate two symbolic reasoning tasks: last letter concatenation (e.g., the input is ``Elon Musk'' and the output should be ``nk''), and Coinflip (e.g., a coin is heads-up, after a few flips is the coin still heads-up?) from~\citet{wei2022chain}.
% \end{itemize}

\begin{figure*}[!t]
\centering
\includegraphics[width=0.27\linewidth]{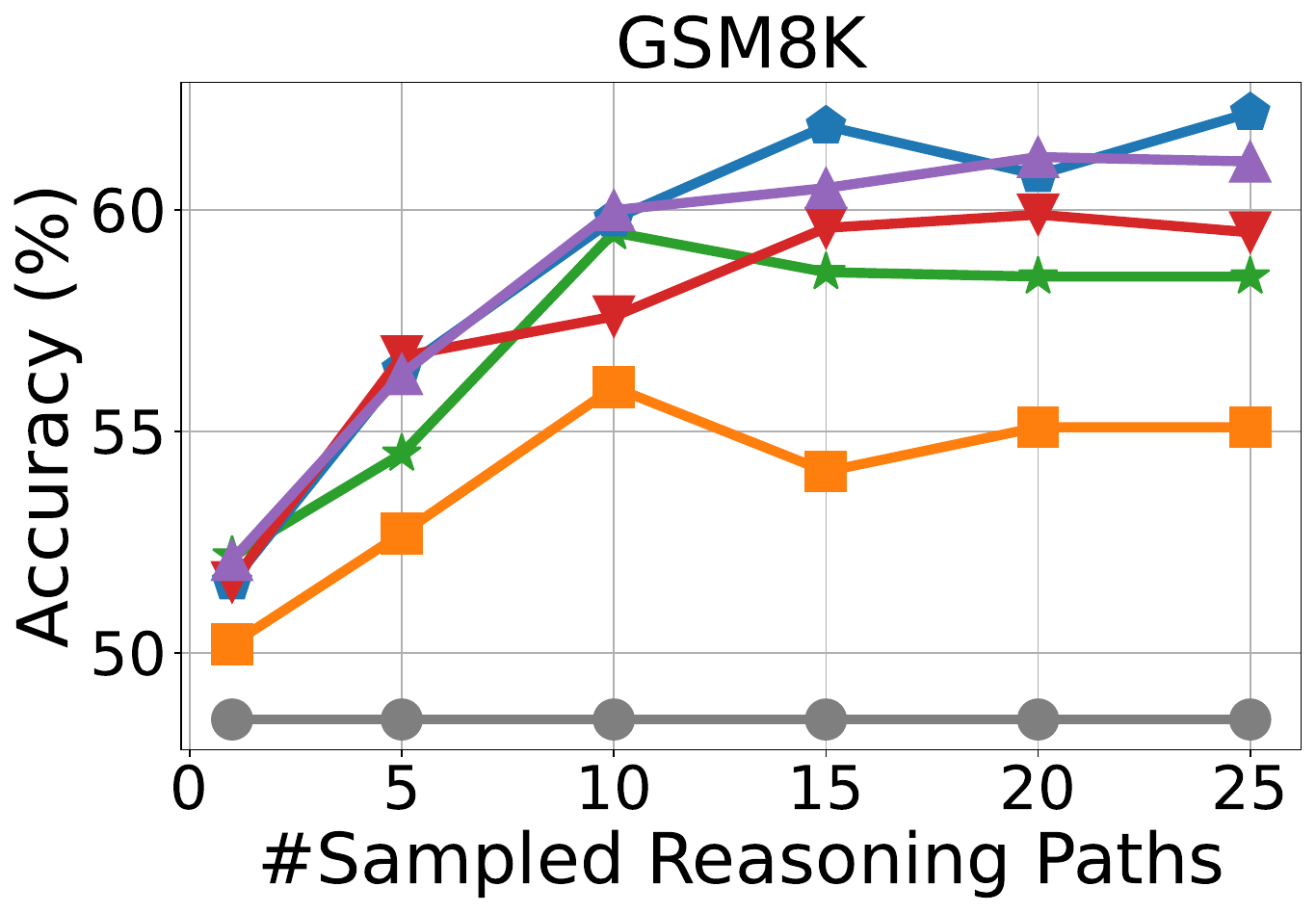}
    \includegraphics[width=0.27\linewidth]{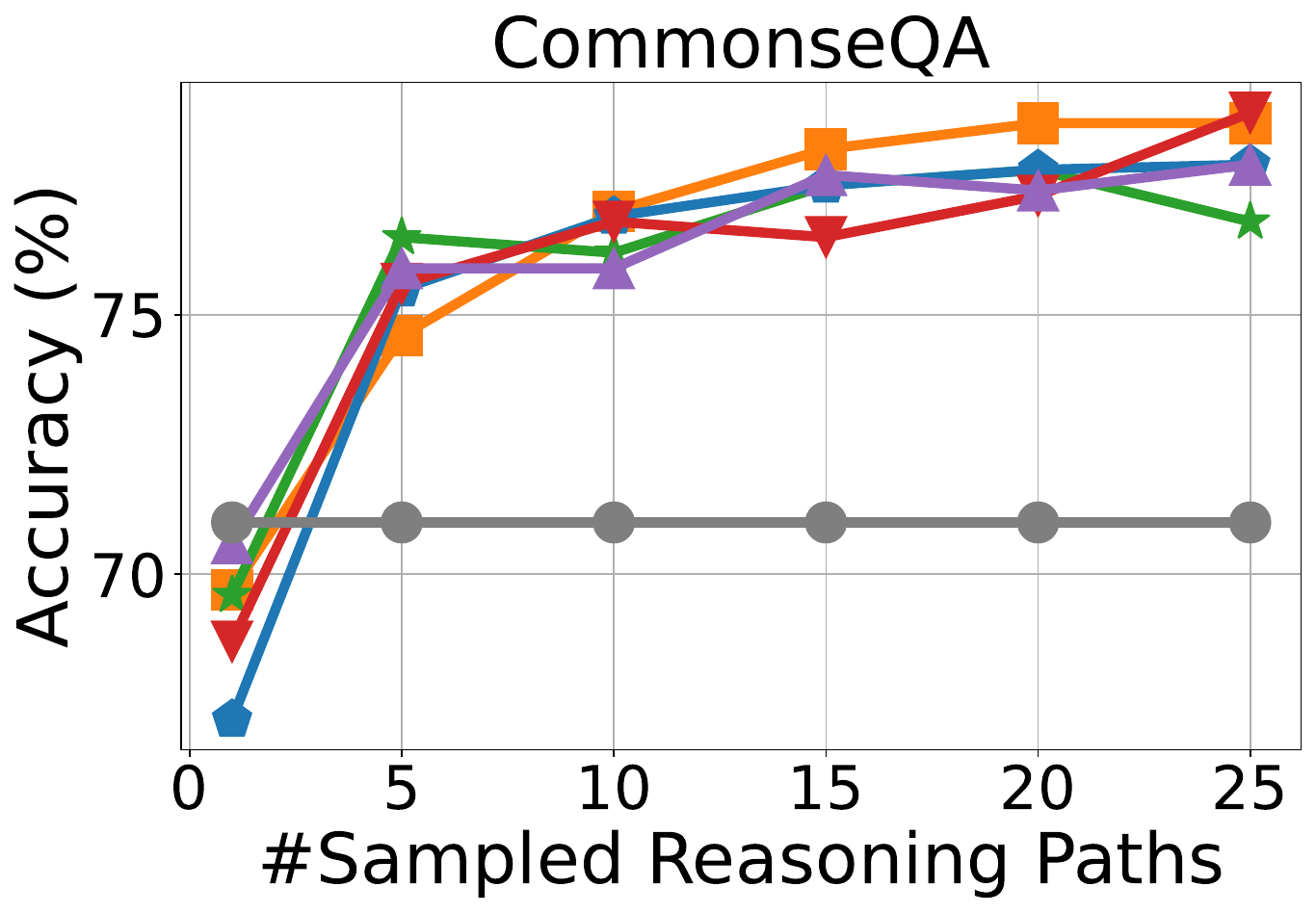}
    \includegraphics[width=0.415\linewidth]{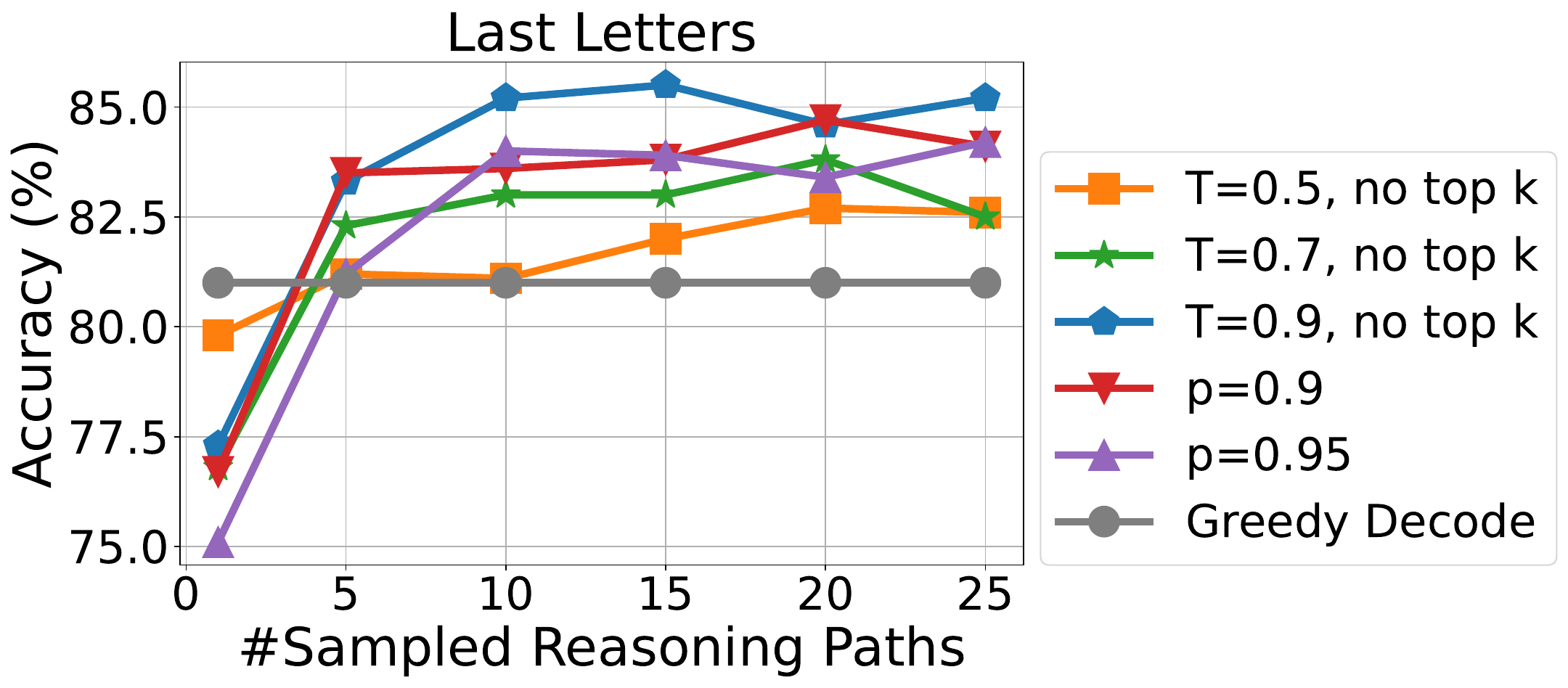}
    \caption{Self-agreement significantly improves accuracy over zero-shot CoT with greedy decoding (grey) on the first scenario across arithmetic, commonsense and symbolic reasoning tasks, over GPT-3.5-turbo. Sampling a higher number of diverse reasoning paths consistently improves reasoning accuracy. In addition, self-agreement is robust to reasoning paths with various diversity across arithmetic, commonsense and symbolic reasoning tasks.
    }
    \label{fig:fig4}
\end{figure*}
\noindent\textbf{Language Models} We evaluate self-agreement over two transformer-based language models on the first and second scenarios, and four transformer-based language models on the third scenario. For the first and second scenarios, we consider the following language models: (i) GPT-3.5-turbo. We use the public \textit{gpt-3.5-turbo} version of GPT-3.5 from OpenAI API;\footnote{\url{https://openai.com/blog/openai-api}} (ii) Llama-2-13B-Chat~\cite{touvron2023llama} with 13-billion parameters, optimized for dialogue use cases via  alignment techniques. Llama-2-Chat models are completely open-sourced\footnote{Model checkpoints and inference code are available at~\url{https://github.com/facebookresearch/llama}.} and have similar performance compared with GPT-3~\cite{brown2020language}. For the third scenario, we consider the following language models: (i) LaMDA-137B~\cite{thoppilan2022lamda} with 137-billion parameters, pre-trained on a mixture of web documents, dialog data and Wikipedia; (ii) PaLM-540B~\cite{chowdhery2022palm} with 540-billion parameters, pre-trained on a high quality corpus of 780 billion tokens; (iii) Minerva-540B~\cite{lewkowycz2022solving} with 540-billion parameters, pretrained on
general natural language data and further trained on technical content; (iv) GPT-3.5-turbo. 
% We use the public \textit{gpt-3.5-turbo} version of GPT-3.5.
% \begin{itemize}
%     \item \textbf{Close-sourced Language Models.} We first report the existing state-of-the-art (SOTA) performance in fine-tuning field. Next, we consider the following language models for prompting field: (1) LaMDA~\cite{thoppilan2022lamda}: a 137B model; (2) PaLM: a 540B model; (3) Minerva: a 540B model; (4) GPT-3.5: GPT3.5-turbo from~\citet{brown2020language}.
%     \item  \textbf{Open-sourced Language Models.} We consider the following language models for prompting filed: Llama 2-Chat with 13B and 70B sizes, respectively.
% \end{itemize}
% We perform all experiments in the few-shot setting, without training or fine-tuning the language models. We use the same prompts from~\citet{wei2022chain} for a fair comparison.

\noindent\textbf{Baselines} 
% As shown in Figure~\ref{fig:fig2}, there are three major cases covering almost all possible situations. 
As shown in Figure~\ref{fig:fig2}, there are three major scenarios covering almost all possible situations. In the first scenario, we compare self-agreement with 2 baselines: (i) Zero-Shot CoT~\cite{kojima2022large}; (ii) Self-Consistency~\cite{wang2022self}. In the second scenario, we compare self-agreement with 3 baselines: (i) Mixed-Few-Shot CoT. To adapt few-shot CoT to such scenario, we randomly collects one demonstration from each reasoning task used in~\citet{wei2022chain} and then leverage the mixed demonstrations for all input questions. (ii) Self-Consistency~\cite{wang2022self}; (iii) Multi-Agents (Debate)~\cite{du2023improving}. In the third scenario, we compare self-agreement with 3 baselines: (i) Few-Shot CoT~\cite{wei2022chain}; (ii) Self-Consistency~\cite{wang2022self}; (iii) USC~\cite{chen2023universal}.

\noindent\textbf{Sampling Schemes} To sample diverse reasoning paths in the \textit{ask k times} stage, we follow the same settings of~\citet{wang2022self}. In particular, we apply temperature sampling with $T$ = 0.5 and truncated at the top-$k$ ($k$ = 40) tokens with the highest probability for Llama-2-Chat models with various sizes. For GPT-3.5-turbo, we use $T$ = 0.7 without top-$k$ truncation.

\noindent\textbf{Evaluation Metrics} For all datasets, we use accuracy to evaluate the model's reasoning performance.

\subsection{Main Results}
\label{sec:mar}

We report the results of self-agreement averaged over 5 runs, where we sample 20 outputs from the model in each run.
% independently from.\footnote{To be economical, we sample 100 outputs from the model. For GPT-3.5-turbo, we sample 40 outputs independently from 100 outputs for each run. For Llama-2-Chat models, we sample 25 outputs independently from 100 outputs for each run due to the limits of the maximum context length.} 
% We compare two prompting schemes with greedy decoding: (1) standard prompting. We reuse a few examples constructed originally with~\citet{wei2022chain} without intermediate reasoning steps; (2) CoT prompting. We reuse a few examples constructed originally with~\citet{wei2022chain}.

% \noindent\textbf{Performance of Self-Agreement on the First Scenario} 
The results on the first scenario with six datasets from three categories are shown in Table~\ref{table:t1}. Self-agreement improves the reasoning performance in all reasoning tasks across all two language models over zero-shot CoT without requiring the answer formats to be similar, while self-consistency is not applicable in this scenario as the reasoning paths have no trigger (i.e., \textit{The answer is}). 
% Furthermore, self-agreement clearly outperforms multi-agents (debate) in almost all reasoning tasks across all two language models, despite the fact that self-agreement is simpler.
% , unsupervised, effective and agnostic of both task and scenario. 
More surprisingly, the gains brought by our method are almost the same across different language models' scales,
% , while multi-agents (debate) attains more competitive performance compared with zero-shot CoT when using GPT-3.5-turbo as agents, 
further demonstrating the general effectiveness of our method. For example, we see +5.8\% average absolute accuracy improvement over Llama-2-13B-Chat and +5.3\% for GPT-3.5-turbo when employing self-agreement.
We can also discern a generally consistent performance trend in the second scenario, mirroring that of the first scenario. Self-agreement improves the reasoning performance (average absolute accuracy +5.5\% for GPT-3.5-turbo and +6.2\% for Llama-2-13B-Chat) across all two language models over mixed-few-shot CoT, while self-consistency performs even worse (average absolute accuracy -1.2\% for GPT-3.5-turbo and -0.5\% for Llama-2-13B-Chat) than mixed-few-shot CoT. Because determining whether two strings represent \textbf{the same final answer} is not applicable for it when the type of input questions is unknown. We provide further analysis in Appendix~\ref{sec:aexp3}. It can also be seen that multi-agents (debate) attains more competitive performance compared with zero-shot CoT when using GPT-3.5-turbo as agents. We qualitatively find that it is more difficult for models of small scale to arrive at the correct answer as the debate progresses when facing cases where all the agents initially make incorrect predictions (see Table~\ref{table:t9} in the Appendix). The underlying reason is that the process of debating is an emergent ability of model scale~\cite{wei2022emergent}. That is, stronger models can better understand the debating rules and refine other agents' incorrect outputs for further improvement.
% Because determining whether two strings represent \textbf{the same final answer} is not applicable for it when the type of input questions is unknown. For instance, \textit{6 hours} and \textit{6} extracted from outputs 2, 3 of Figure~\ref{fig:fig2} respectively represent two different answers when employing self-consistency in this scenario, since we can only extract the strings between the trigger (e.g., ``The answer is'') and full stop (e.g., ``.''). 
% More surprisingly, the gains are almost the same across different language models' scales, which is the same as the conclusions on the first scenario.

% \noindent\textbf{Performance of Self-Agreement on the Third Scenario}
\begin{table}[t]
  \centering
  \small
  \resizebox{1.0\linewidth}{!}{
  \begin{tabular}{lllccc}
      \toprule
      \bf Model & & & \bf GSM8K & \bf CSQA & \bf Letter\\
      \midrule
    \multirow{4}{*}{GPT-3.5-turbo} & \multicolumn{2}{l}{Zero-Shot CoT} & 48.5 & 71.0 & 81.0\\
    & \multicolumn{2}{l}{Self-Consistency} & N/A & N/A & N/A \\
    % & \multicolumn{2}{l}{Multi-Agents (Debate)} & 80.2 & 69.0 & 61.3 \\
    & \multicolumn{2}{l}{Self-Agreement (w/ original prompt)} & 58.5 & 77.8 & 83.8 \\
    & \multicolumn{2}{l}{Self-Agreement (w/ modified prompt)} & 56.7 & 77.5 & 83.4 \\ 
      \bottomrule
  \end{tabular}
  }
  \caption{
  Self-agreement works with different prompts.
  }
  \label{table:t3}
\end{table}
\begin{figure*}[!t]
\centering
\includegraphics[width=0.285\linewidth]{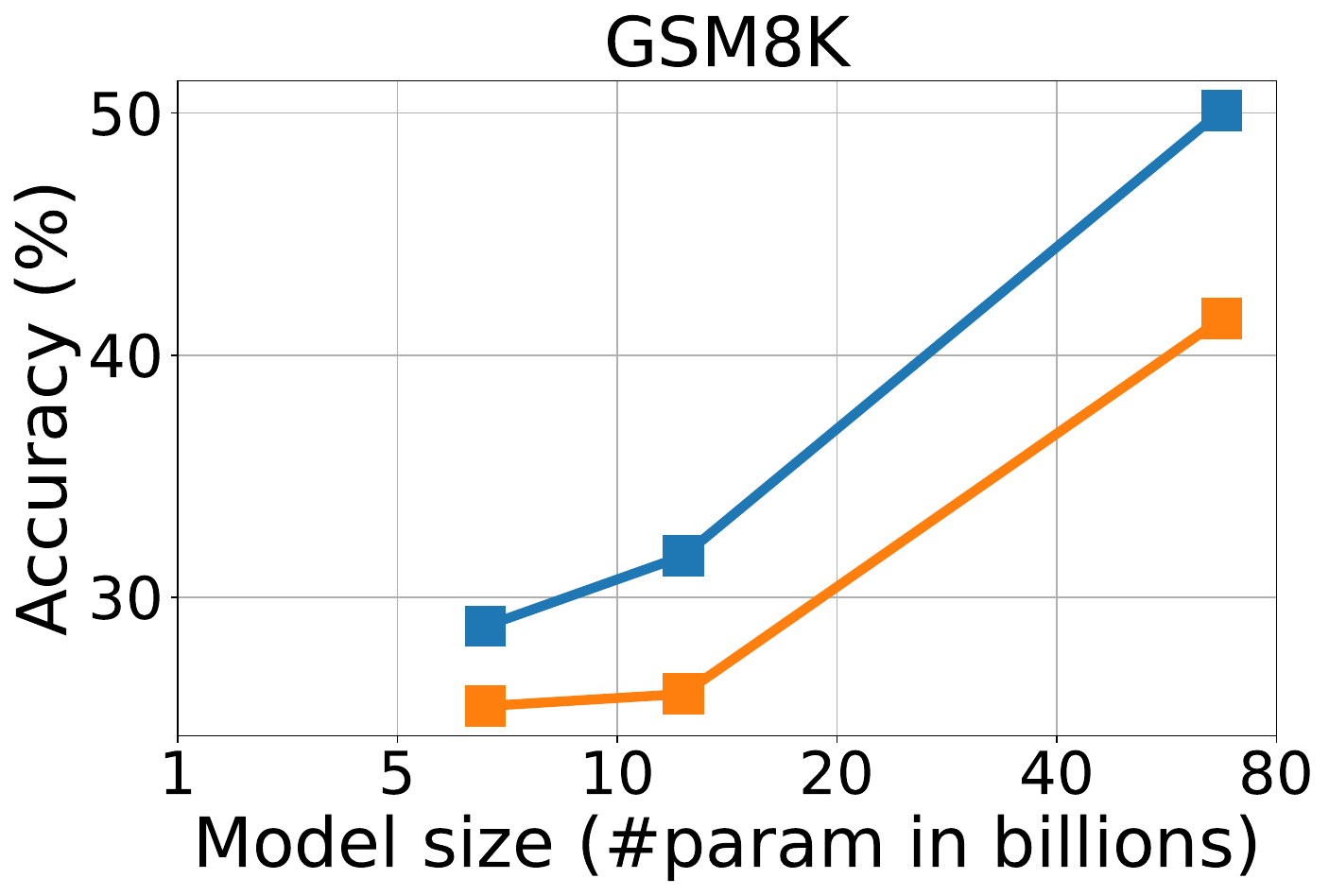}
    \includegraphics[width=0.285\linewidth]{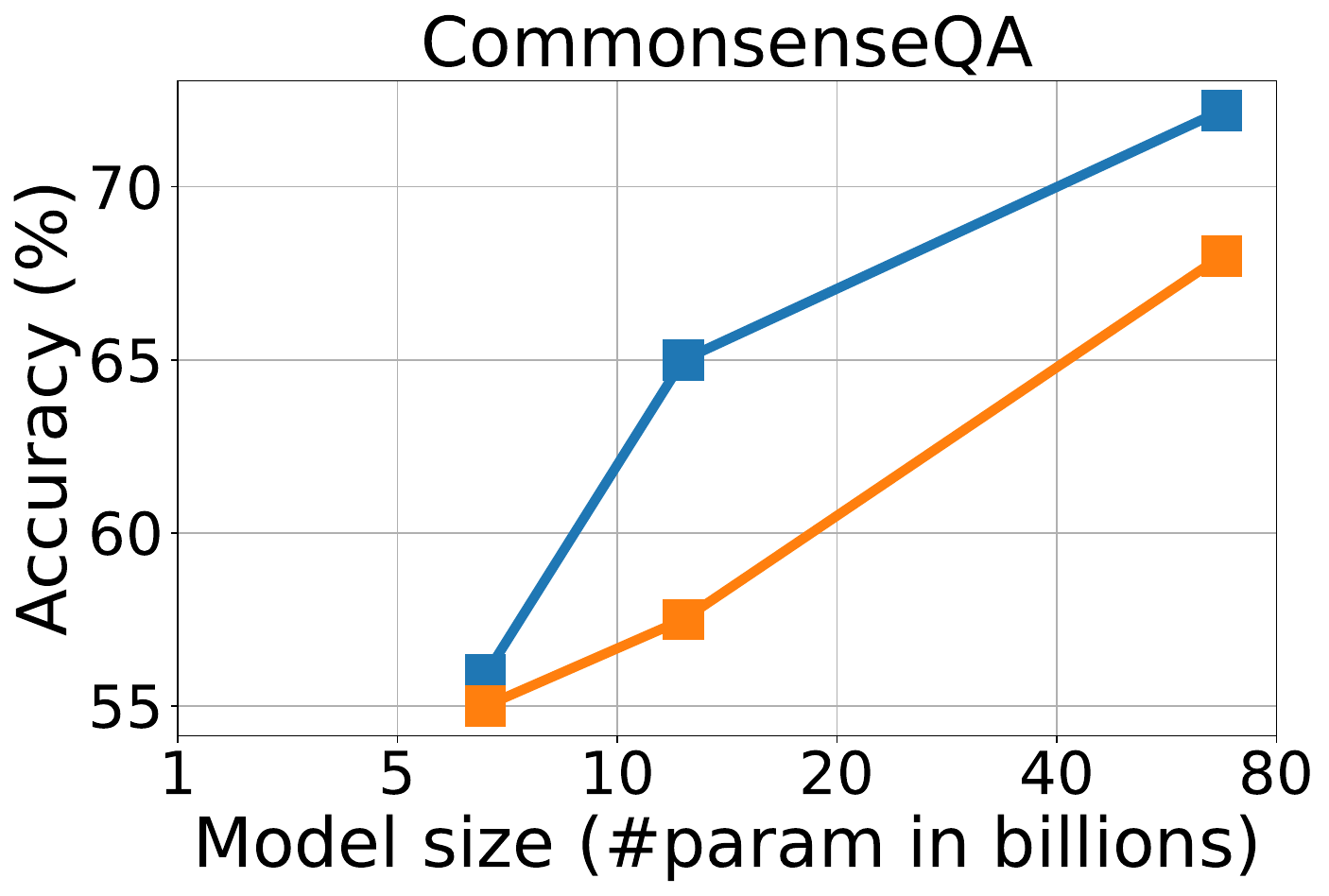}
    \includegraphics[width=0.40\linewidth]{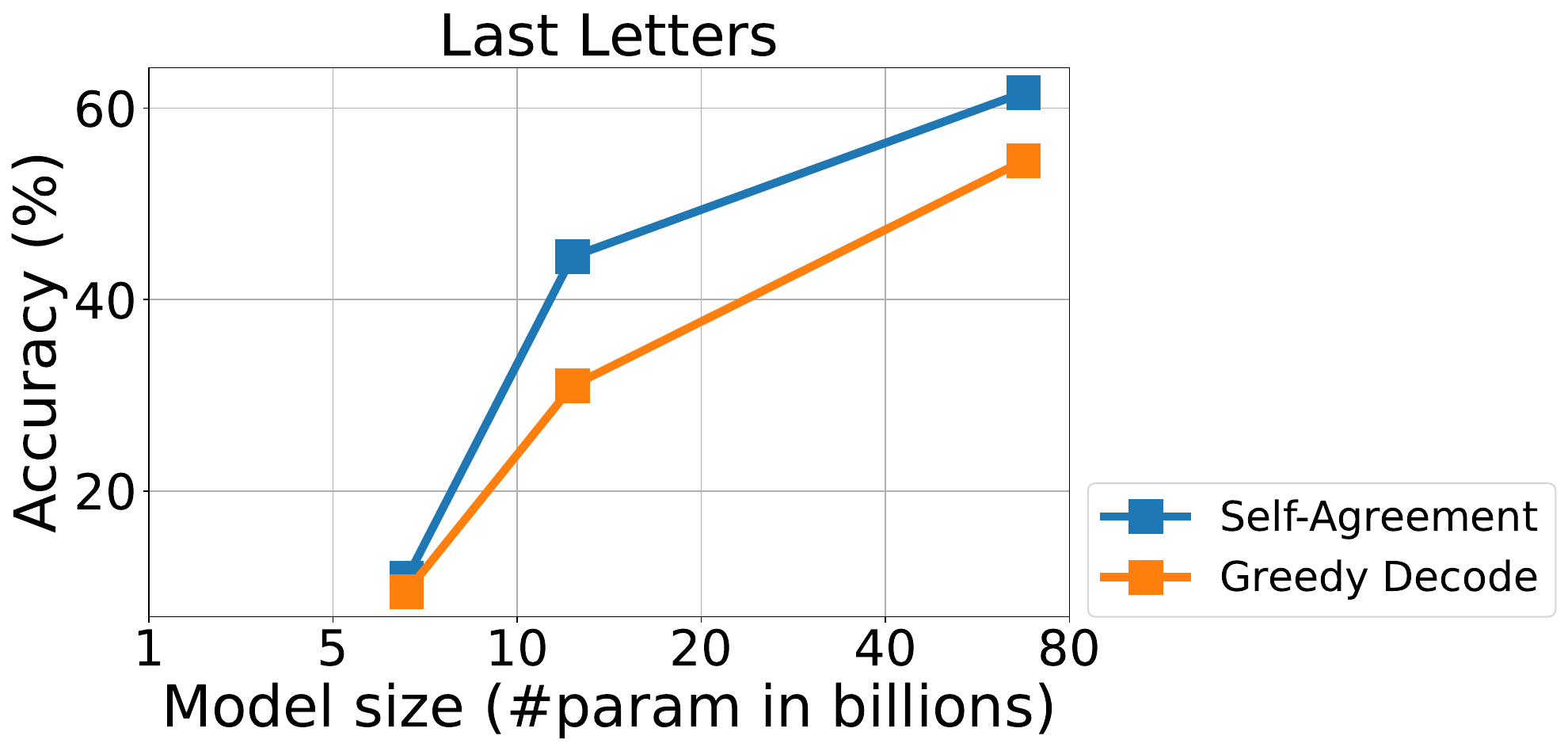}
    \caption{Self-agreement improves performance across different language model scales.}
    \label{fig:fig5}
\end{figure*}
Table~\ref{table:t2} summarizes accuracy of our method, few-shot CoT, self-consistency and USC for each dataset. Similarly, self-agreement yields large gains over few-shot CoT for all reasoning tasks. Meanwhile, self-agreement achieves almost the same superior performance as self-consistency across almost all reasoning tasks, which it does not need answer parsing to perform the voting. On the contrary, the performance of USC is far inferior to self-consistency and self-agreement, indicating that self-agreement has better expandability and can really approach the performance ceiling of self-consistency while USC cannot. This is also demonstrated in~\citet{chen2023universal}. We consider there is an inescapable gap between performance and generalization. We provide detailed discussion with USC in Appendix~\ref{sec:dis}. We also find that self-agreement has larger performance gains for more-complicated problems. For instance, in Arithmetic Reasoning, for GSM8K (the dataset with the lowest baseline performance), the performance gain (+12.4) is more than tripled for other tasks. This is analogous to the intuition self-agreement leveraged that complex reasoning tasks typically admit multiple reasoning paths that reach a correct answer. The more that deliberate thinking and analysis is required for a problem, the greater the diversity of reasoning paths that can recover the answer.
% We focus on how to solve open-ended reasoning tasks, not open-ended generation tasks. Because the ability of LLMs to challenging tasks (e.g., arithmetic, commonsense and symbolic reasoning) is often seen as a limitation, which is difficult to be overcome solely by scaling up the size of LLMs .

% However, we notice that the gap between them is relatively large on the Last Letters dataset. We provide further analysis in Appendix~\ref{sec:aexp2}.
% does not positively impact performance for all datasets relative to self-consistency. We qualitatively find that rule-based processing can not evaluate the majority voted answer generated by self-agreement effectively. We also provide a case in Table~\ref{table:t10}.

% The single-task scenario we defined is both the type of questions and answer format are known, which is the commonly assumed situation in previous ensemble-optimization studies.

% Self-agreement improves the reasoning performance across two language models significantly over chain-of-thought prompting.

\subsection{Analysis}
\label{sec:analysis}

We conduct a number of additional experiments to analyze different aspects of our approach. For all experiments, we use GPT-3.5-turbo, and conduct experiments on the first scenario as it is the most relevant to the real-world scenario, unless otherwise specified. We only mix GSM8K, CommonsenseQA and Last Letters selected from each category to simulate the first scenario.
% We only use GSM8K, CommonsenseQA and Last Letters selected from each category, and then mix them to simulate the first scenario.

\noindent\textbf{Effects of the Number of Sampled Reasoning Paths} We argue that the most agreed answer selected from multiple reasoning paths is likely to be the correct answer. So, we are curios about the effect of the number of sampled reasoning paths. 
% In this experiment, we investigate its influence in GSM8K, CommonsenseQA and Last Letters on the first scenario. 
As shown in Figure~\ref{fig:fig4}, we observe that the performance first increases sharply and then slows down as the number of sampled reasoning paths increases. This is why we sample 20 outputs for all scenarios and tasks. In addition, sampling a higher number of reasoning paths leads to a consistently better performance, but performance nearly converges at the number of 5 or 10. This is consistent with our statements that assessing whether or not a person really knows how to solve a question should depend on a small number of reasoning paths generated by themself, since a person might acquire the wrong answer due to
carelessness or other factors.

\noindent\textbf{Self-Agreement is Robust to Reasoning Paths with Different Diversity and Scaling} Sampling strategies with different hyperparameters denote generated reasoning paths with various diversity. So, we are curios about whether self-agreement is robust to reasoning paths with various diversity. To show the effect of reasoning paths with various diversity, we conduct the experiment by varying \textit{T} in  temperature sampling~\cite{ficler2017controlling} and \textit{p} in nucleus sampling~\cite{holtzman2019curious}, over GPT-3.5-turbo. As shown in Figure~\ref{fig:fig4}, we can see that self-agreement gradually improves reasoning performance regardless of the diversity of reasoning paths as the number of reasoning paths increases. It also suggests that self-agreement is robust to reasoning paths with various diversity. Figure~\ref{fig:fig5} shows that self-agreement robustly improves performance in GSM8K, CommonsenseQA and Last Letters across all scales for the Llama-2-Chat model series (i.e., 7B, 13B and 70B). It is worth noting that the gain when used with Llama-2-7B-Chat is relatively low due to the fact that certain abilities only emerge when the model reaches a sufficient scale~\cite{brown2020language}.

\noindent\textbf{Effects of Different Self-Agreement Prompts} We further carry out experiments to examine the influence of the prompt used in \textit{ask one more time} stage of self-agreement. Specifically, we modify the original prompt by first translating it into another language and then back again to construct the modified prompt. The original and modified prompts are given in Table~\ref{table:t5} and~\ref{table:t6} in the Appendix. As shown in Table~\ref{table:t3}, we can observe self-agreement (w/ modified prompt) achieves almost the same performance, indicating that self-agreement is robust to the prompts with the same meaning.
% used in \textit{ask one more time} stage.

\section{Conclusion}
\label{sec:conclu}

In this work, we explore a more realistic setting with significant application values in ensemble-optimization studies, as shown in Figure~\ref{fig:fig2}.
% , where the type of input questions is unknown and the answer format of reasoning paths may be known or unknown. 
% To address such scenarios while maintain the superior performance in the third scenario, 
To this end, we propose self-agreement, a simple yet generalizable ensemble-optimization method applying in almost all scenarios. Self-agreement simultaneously achieves remarkable performances on six public reasoning benchmarks and superior generalization capabilities. Our findings encourage the research community to focus on a deeper understanding of the role of language model itself in ensemble-optimization studies, which has favorable generalization abilities and superior performance. We hope this work will open new research possibilities in prompting, language models, ensemble-optimization and multi-step reasoning.

\section*{Limitations}
There are two limitations of our approach. First, the \textit{ask k times} stage of self-agreement incurs more computational cost. We suggest people can try a small number of paths (e.g., 5 or 10) to achieve most of the performance gains while not bringing too much cost, since we have shown that performance nearly converges at the number of 5 or 10. Besides, there are some work investigating how to enable the LLM to run inference in batches~\cite{cheng2023batch,lin2023batchprompt}. As part of future work, one could mix different questions as a batch, and run inference in a batch one time. After that, self-agreement requires only two inferences, thus significantly decreasing computational cost. Second, in the \textit{ask one more time} stage, we modify the input question $\mathbf{x}$ and its multiple reasoning paths $\mathbf{z}_{i}$ into a \textit{prompt} $\mathbf{s}$. This may cause the length of a \textit{prompt} $\mathbf{s}$ exceeds the maximum context length of language models. For example, the maximum context length of Llama-2-Chat models and GPT-3.5-turbo is 4,096 and 8,192 respectively. In practice, as shown in Figure~\ref{fig:fig4}, performance nearly converges at the number of 5 or 10. So, the maximum context length of language models is likely to be enough for self-agreement in most situations. Meanwhile, there are also some work focusing on how to increase the context length of language models~\cite{longchat2023,xiong2023effective}. Therefore, in the future, the limitations of the context length will become smaller and smaller.

% Entries for the entire Anthology, followed by custom entries
\bibliography{anthology,custom}
\bibliographystyle{acl_natbib}

\appendix
\newpage

\section{Experimental Details}
\label{sec:app_ed}

\subsection{Tasks and Datasets}
\label{sec:app_td}

We evaluate self-agreement on six public reasoning benchmarks for a fair comparison with existing methods, covering arithmetic, commonsense and symbolic reasoning tasks. The detailed statistics of the datasets are shown in Table~\ref{table:t4}. 

\begin{table}[htp]
\centering
\small
% \vspace{2.8mm}
% \setlength{\tabcolsep}{13.6pt}
\resizebox{1.0\linewidth}{!}{
\begin{tabular}{lcrr}
\toprule
\bf Dataset & \bf Number of samples & \bf Answer Format & \bf Licence\\
\midrule
MultiArith &600  &Number & Unspecified \\
GSM8K  &1,319 &Number & MIT License \\
SVAMP  &1,000 &Number & MIT License \\
CSQA & 1,221  &Multiple choice & Unspecified \\
Last Letters &500 & String & Unspecified \\
ARC-c & 1,172 &Multiple choice & CC BY-SA\\
% Coin Flip  &500 & Yes or No & Unspecified \\
\bottomrule
\end{tabular}
}
\caption{Detailed dataset description.}
\label{table:t4}
\end{table}

\subsection{Baseline Methods}
\label{sec:app_bm}

We introduce the baseline methods in detail.

\paragraph{Few-Shot CoT} Few-Shot CoT~\cite{wei2022chain} employs several additional templated demonstrations as: ``Q: \texttt{q}. A: \texttt{(r, a)}.'' before the input question, where \texttt{q}, \texttt{r} and \texttt{a} are manually crafted questions, rationales and final answers.

\paragraph{Zero-Shot CoT} Zero-Shot CoT~\cite{kojima2022large} simply inserts the prompt ``Let’s think step by step'' after the input question.

\paragraph{Self-Consistency} Self-Consistency~\cite{wang2022self} first samples multiple reasoning paths by using few-shot CoT, and then selects the most consistent answer by marginalizing out the sampled reasoning paths.

\paragraph{Multi-Agents (Debate)} Multi-Agents (Debate)~\cite{du2023improving} takes language models as multi-agents to propose and debate their individual responses and reasoning processes over multiple rounds
to arrive at a common final answer.

\paragraph{USC} USC~\cite{chen2023universal} leverages LLMs themselves to select the most consistent answer among multiple candidates.

\subsection{Implementation Details}
\label{sec:app_id}

For all language models we evaluated, we perform prompting-based inference only. For GPT-3.5-turbo, we use the public version \textit{gpt-3.5-turbo} of GPT-3.5 from OpenAI API. For Llama-2-Chat models, we use the open-sourced library, and run experiments on NVIDIA Tesla A800 (8x8 configuration, 80G).

For GPT-3.5-turbo, we follow the same experimental settings of~\citet{kojima2022large} and set 128, 256 max tokens for \textit{ask k times} and \textit{ask one more time} phases respectively, without frequency penalty or presence penalty. For Llama-2-Chat models, we set 128 max tokens for \textit{ask k times} without frequency penalty or presence penalty. In the \textit{ask one more time} stage, we continue to generate tokens until the stop token (e.g., ``</s>'') is generated. We use greedy decoding across all the models in \textit{ask one more time} phase. Our implementation is in PyTorch~\cite{paszke2019pytorch}.

\begin{table}[t]
  \centering
  \small
  \resizebox{1.0\linewidth}{!}{
  \begin{tabular}{@{}ll  @{\hspace{2pt}} ccc}
      \toprule
      \bf Model & & \multicolumn{1}{c}{\textit{Arithmetic}} & \multicolumn{1}{c}{\textit{Commonsense}} & \multicolumn{1}{c}{\textit{Symbolic}} \\
      \cmidrule(r){3-3}
      \cmidrule(r){4-4}
      \cmidrule(r){5-5}
       & & \bf GSM8K & \bf CSQA & \bf Letter \\\midrule
       \multicolumn{5}{c}{\textit{Both the type of questions and the answer format are unknown (\textbf{the first scenario})}} \\
       \midrule
       \multirow{12}{*}{GPT-3.5-turbo} &\multicolumn{4}{c}{\textbf{200} randomly selected examples} \\
       \cmidrule(r){2-5}
       & Zero-Shot CoT & 48.5 & 71.0 & 81.0 \\ 
                  & Self-Agreement & 58.5 (\blue{+10.0}) &  77.8 (\blue{+6.8}) & 83.8 (\blue{+2.8}) \\
                  \cmidrule(r){2-5}
                  &\multicolumn{4}{c}{\textbf{400} randomly selected examples} \\
                  \cmidrule(r){2-5}
                  & Zero-Shot CoT & 48.8 & 70.5 & 77.8 \\ 
                  & Self-Agreement & 54.3 (\blue{+5.5}) &  77.1 (\blue{+6.6}) & 80.4 (\blue{+3.3}) \\
                  \cmidrule(r){2-5}
                  &\multicolumn{4}{c}{\textbf{600} randomly selected examples} \\
                  \cmidrule(r){2-5}
                  & Zero-Shot CoT & 47.8 & 69.6 & 79.6 \\ 
                  & Self-Agreement & 54.0 (\blue{+6.2}) &  77.1 (\blue{+7.5}) & 81.9 (\blue{+2.3}) \\
      \bottomrule
  \end{tabular}
  }
  \caption{
  Self-agreement, when applied on GPT-3.5-turbo. Our performance gain (\blue{\textbf{+blue}}) is computed over the zero-shot CoT~\cite{kojima2022large}, which is our primary baseline.
  Our method substantially increases the performance over zero-shot CoT. 
  % The numbers in brackets denote the number of randomly selected samples from GSM8K, CommonsenseQA and Last Letters.
  }
  \label{table:t11}
\end{table}
Here, we describe the implementation details of baseline methods in different scenarios. In the first scenario, to adapt zero-shot CoT to such scenario, we use the general prompt (i.e., The answer is) for answer extraction. For all ensemble-optimization methods, we employ zero-shot CoT~\cite{kojima2022large} without 2nd answer extraction phase to simulate the reasoning paths with no trigger. Self-consistency is not applicable in this scenario as the reasoning paths have no trigger. In the second scenario, to adapt few-shot CoT to such scenario, we randomly select one demonstration from each reasoning task used in~\citet{wei2022chain} and then leverage the mixed demonstrations for all input questions (called mixed-few-shot CoT). The demonstrations are given in Table~\ref{table:t7}, and other experimental settings are the same as~\citet{wei2022chain}. To adapt self-consistency to such scenario, we extract the strings between the trigger (i.e., ``The answer is'') and full stop (i.e., ``.'') of the reasoning paths.
% as $\mathbf{a}_{i}$, and then selects the optimal answer among the final answer set by taking a majority vote over $\mathbf{a}_{i}$. 
For example, as shown in Figure~\ref{fig:fig2}, we obtain \textit{6 hours} extracted from the output 2. To adapt multi-agents (debate)~\cite{du2023improving} to such scenario, we modify the prompt used in~\citet{du2023improving} for GSM8K to make it applying in almost all scenarios, since~\citet{du2023improving} have designed a specific prompt for each task, which is not a generalizable method. The modified prompt is given in Table~\ref{table:t8}. Other experimental settings are the same as~\citet{du2023improving}. In the third scenario, we use the same prompts and experiment settings used in~\citet{wei2022chain} to employ few-shot CoT, the same experimental settings of~\citet{wang2022self} to employ self-consistency, and the same experimental settings of~\citet{chen2023universal} to employ USC.

\section{Additional Experiments}
\label{sec:aexp}

\subsection{Are 200 Randomly Selected Examples Enough?}
\label{sec:aexp1}

As mentioned in Section~\ref{sec:expset}, to be economical, we only select 200 examples from each public reasoning benchmark randomly and then mix them to simulate situations of the unknown type of input questions. To further validate the soundness of experimental results in the first scenario provided in Table~\ref{table:t1}, we conduct experiments on GSM8K, CommonsenseQA and Last Letters by varying the number of randomly selected examples (i.e., 200, 400 and 600).\footnote{Note that the test set of Last Letters dataset only contains 500 examples. Thus, we use all examples when randomly selected 600 examples from each reasoning task.} As shown in Table~\ref{table:t11}, the performance fluctuates slightly with different number of randomly selected examples from each reasoning task over zero-shot CoT and self-agreement. Overall speaking, self-agreement improves the reasoning performance in all reasoning tasks over zero-shot CoT.

\subsection{Further Analysis on Self-Agreement vs. Self-Consistency}
\label{sec:aexp2}

To have an entirely fair comparison with self-consistency, we conduct experiments on 20 and 40 sampled reasoning paths respectively. 
\begin{figure}[!t]
\centering
\includegraphics[width=0.49\linewidth]{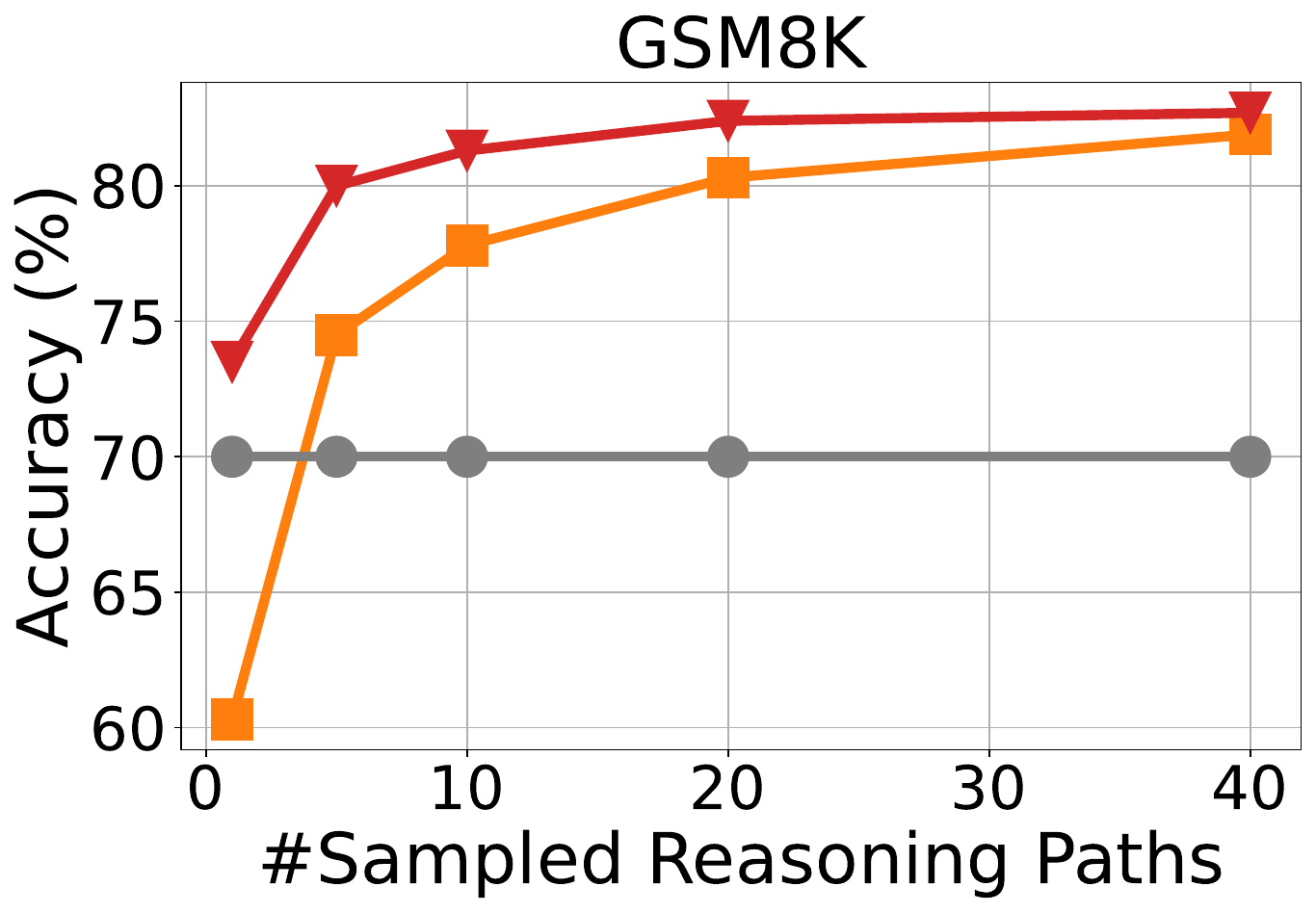}
    \includegraphics[width=0.49\linewidth]{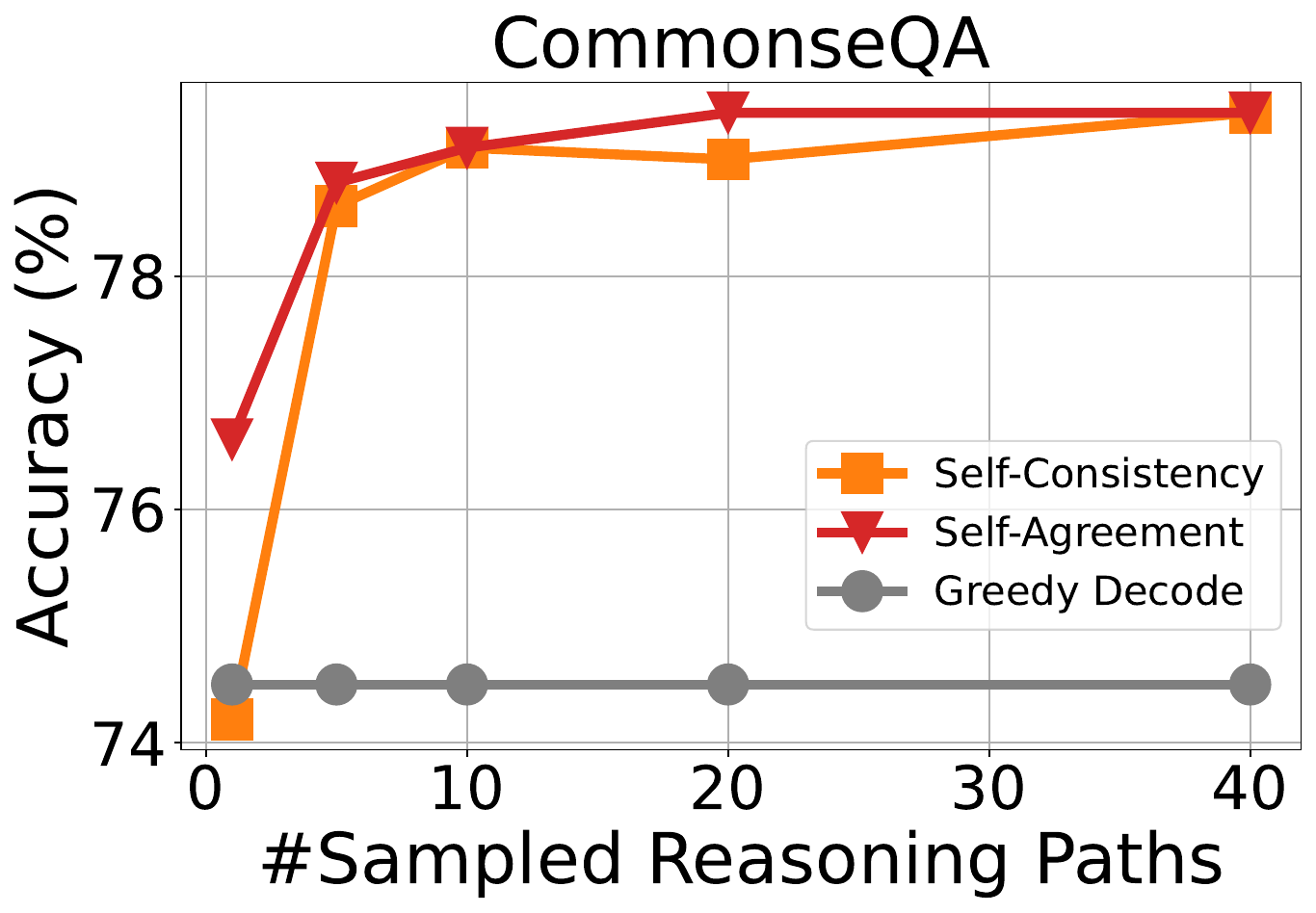}
    \caption{Self-agreement utilizes sampled reasoning paths more efficiently than self-consistency when the number of sampled reasoning paths is less than 20. Sampling a enough high number of reasoning paths achieves essentially the same performance for self-consistency and self-agreement.}
    \label{fig:fig6}
\end{figure}
As shown in Table~\ref{table:t12}, it is interesting to note that when the number of sampled reasoning paths is 20, self-agreement achieves an average performance increase of 0.2 over all reasoning tasks compared to self-consistency, but it instead decreases by 0.3 when the number of sampled reasoning paths is increased to 40. To investigate this influence, we conduct experiments on GSM8K and CommonseQA with self-consistency and self-agreement by varying the number of sampled reasoning paths. As shown in Figure~\ref{fig:fig6}, self-agreement utilizes sampled reasoning paths more efficiently than self-consistency when the number of sampled reasoning paths is less than 20. Sampling a enough high number of reasoning paths achieves essentially the same performance for self-consistency and self-agreement. The underlying reason is that self-agreement is a generalizable answer extractor while self-consistency is a rule-based answer extractor, i.e., only extract the reasoning paths with trigger (e.g., \textit{The answer is}). This causes it to drop reasoning paths without trigger.

\begin{table*}[t]
  \centering
  \small
  \resizebox{0.95\linewidth}{!}{
  \begin{tabular}{@{}ll  @{\hspace{2pt}} cccccc}
      \toprule
      \bf Model & & \multicolumn{3}{c}{\textit{Arithmetic}} & \multicolumn{2}{c}{\textit{Commonsense}} & \multicolumn{1}{c}{\textit{Symbolic}} \\
      \cmidrule(r){3-5}
      \cmidrule(r){6-7}
      \cmidrule(r){8-8}
       & & \bf GSM8K & \bf MultiArith & \bf SVAMP & \bf CSQA & \bf ARC-c  & \bf Letter \\\midrule
       \multicolumn{8}{c}{\textit{Both the type of questions and the answer format are known (\textbf{the third scenario})}} \\
        \midrule
      \multirow{10}{*}{GPT-3.5-turbo} & \multicolumn{7}{c}{\textbf{20} sampled reasoning paths} \\
      \cmidrule(r){2-8} 
                  & Self-Consistency & 80.3 & \bf 99.2 & 85.9 & 79.0 & 87.0 & 81.9 \\ 
                  & USC & 76.8 & 98.2 & 83.5 & 48.9 & 73.9 & 79.4 \\ 
                  & Self-Agreement & 82.4 (\blue{+2.1}) & 99.0 (\blue{-0.2}) & 86.0 (\blue{+0.1}) & \bf 79.4 (\blue{+0.4}) & 86.8 (\blue{-0.2}) & 81.0 (\blue{-0.9}) \\
    \cmidrule(r){2-8}
    & \multicolumn{7}{c}{\textbf{40} sampled reasoning paths} \\
      \cmidrule(r){2-8} 
                  & Self-Consistency & 81.9 & \bf 99.2 & \bf 86.5 & \bf 79.4 & \bf 87.1 & \bf 82.4 \\ 
                  & USC & 76.0 & 98.7 & 82.4 & 63.7 & 79.1 & 73.8 \\ 
                  & Self-Agreement & \bf 82.7 (\blue{+0.8}) & 99.0 (\blue{-0.2}) & 86.2 (\blue{-0.3}) & \bf 79.4 (\blue{+0.0}) & 86.9 (\blue{-0.2}) & 81.5 (\blue{-0.9}) \\
      \bottomrule
  \end{tabular}
  }
  \caption{
  Self-agreement, when applied on GPT-3.5-turbo. Our performance gain (\blue{\textbf{+blue}}) is computed over self-consistency~\cite{wang2022self}. The best performance for each task is shown in bold.
  }
  \label{table:t12}
\end{table*}
\begin{algorithm*}[t]
\definecolor{codeblue}{rgb}{0.25,0.5,0.5}
\lstset{
  backgroundcolor=\color{white},
  basicstyle=\fontsize{7.2pt}{7.2pt}\ttfamily\selectfont,
  columns=fullflexible,
  breaklines=true,
  captionpos=b,
  commentstyle=\fontsize{7.2pt}{7.2pt}\color{codeblue},
  keywordstyle=\fontsize{7.2pt}{7.2pt},
  escapeinside=``,
}
\begin{lstlisting}[language=python]
# answer_cleaning: extract the final answer from a reasoning path to a question that belongs to a known task. We adopt the code in https://github.com/kojima-takeshi188/zero_shot_cot/blob/main/utils.py
# questions: questions
# groundtruth_answers: groundtruth answers to each question
# diverse_answers: multiple sampled reasoning paths to each question
# maj_ans: return the string with the most occurrences
# judge_ans: return True if two answers are the same or False

correct_counts = 0  # the number of correct answers
for i in range(len(questions)):
    pred_lists = [] # collect the final answer extracted from each reasoning path
    groundtruth_answer = groundtruth_answers[i]
    for j in range(len(diverse_answers[i])):
        pred = answer_cleaning(diverse_answers[i][j])  # extract the answer of j reasoning path to i question
        if pred:
            pred_lists.append(pred)
    optimal_answer = maj_ans(pred_lists) # return the string with the most occurrences in pred_list
    if judge_ans(groundtruth_answer, optimal_answer):
        correct_counts += 1
print(correct_counts / len(questions))
\end{lstlisting}
\caption{Pseudocode of Self-Consistency in a Python-like style.}
\label{algo1}
\end{algorithm*}
\subsection{Why Self-Consistency Perform Worse in the Second Scenario?}
\label{sec:aexp3}

To better figure out this question, we need to review the theory and implementations of self-consistency~\cite{wang2022self}. In detail, given a question, self-consistency first sample from the language model's decoder to generate diverse multiple reasoning paths, extract the final answers of sampled reasoning paths to form a answer set $\mathbf{a}_{i}$, and then choose the answer with the most occurrences. The overall process is outlined in Alogorithm~\ref{algo1}. It is clear that the most important step for the self-consistency method is how to extract the answers of reasoning paths exactly, while the \textit{answer\_cleaning} function is to perform different extraction rules according to different types of input questions. For instance, we extract arabic numerals for GSM8K, but (A - E) for CommonseQA. To adapt self-consistency to this scenario, we extract the strings between the trigger (e.g., ``The answer is'') and full stop (e.g., ``.'') as the final answers. This is why self-consistency even perform worse than the baseline mixed-few-shot CoT in this scenario. Because determining whether two strings represent \textbf{the same final answer} is not applicable for it when the type of input questions is unknown. For instance, \textit{6 hours} and \textit{6} extracted from outputs 2, 3 of Figure~\ref{fig:fig2} respectively represent two different answers when employing self-consistency in this scenario.

% \subsection{Future Work of Self-Agreement}
% \label{sec:aexp3}

% 这里主要可以讲self-agreement的进一步应用、和未来畅想的方向，比如，跳脱推理任务之外。sa提供了一个通用的方法解决通用场景下的推理任务。但是，原则上，你可以设计合适的提取答案原则而后进行大多数投票，或者设计一个合适的比较指标将sa应用于其他任务。

\subsection{Further Exploration of Self-Agreement on Open-ended Reasoning Task}
\label{sec:aexp4}

As shown in Figure~\ref{fig:fig2}, we only categorize the types of questions into known and unknown from a specific angle, i.e., whether we can identify in advance that the question belongs to a particular task. However, we wonder if this is really the case in real-world scenarios. In practice, there exists another angle of its division. In detail, a reasoning path to a question has multiple different sub-answers consisting of the final answer. For instance, a question ``Axel has 50 silver pesos and 80 gold pesos ..., How many pesos does Anna have? What's the total number of pesos they have together?'' and its groundtruth answer ``Anna has 140 pesos and they have a total of 270 pesos together.'', or a question ``What is Beijing? Answer Choices: (A) a city (B) the capital of China (C) an island'' and its groundtruth answer ``The answer is A and B.''.
% , e.g., single-variable numerical or multi-variable numerical questions, single-choice or multiple-choice questions, etc. 
Sadly, as far as we know, there is no such publicly available reasoning dataset to test the effectiveness of our proposed self-agreement. We guess that there are two reasons. First, constructing such a dataset is more labor intensive. Second, it is difficult to assess the accuracy of the final answers when solving such questions and most likely requires human evaluation.

Additionally, we consider that 
% almost all previous ensemble-optimization studies fail to attain competitive performance even compared with baselines when encountering this type of input questions, yet 
our method still can achieve substantially better performance over baselines when encountering this type of input questions. To answer this, we explain the nature of most ensemble-optimization work and self-agreement in terms of definitions and experiments. 

\begin{table}[!t]
    \centering
    \small
    \resizebox{1.0\linewidth}{!}{
    \begin{tabular}{llc}
     \toprule
     \bf Model & & \bf GSM8K-Multi \\
     \midrule
     \multirow{3}{*}{GPT-3.5-turbo} & Zero-Shot CoT & 59.7 \\
     & Self-Consistency & N/A \\
     % & Multi-Agents (Debate) & 27.9 \\
     & Self-Agreement & \bf 64.3 (+4.6) \\
     \bottomrule
    \end{tabular}
    }
    \caption{Self-agreement achieves better performance over all strong baselines on GSM8K-Multi.}
    \label{table:t21}
\end{table}
From the perspective of definitions, we view most ensemble-optimization work as different implementations of \textit{the majority voted idea}.\footnote{Details refer to corresponding papers.} For example, self-consistency~\cite{wang2022self} takes the majority vote over sampled reasoning paths, while multi-agents (debate)~\cite{du2023improving} obtains the common final answer over multiple agents' outputs. It is clear that the essence of them is based on the majority voted design, and our work is in lines with it, but we explore how to allow language model itself to achieve the overall procedure. So, the main difference between them lies in two major steps consisting of the process of the majority vote, i.e., answer extraction and answer comparison. To be specific, self-consistency and multi-agents (debate) are rule-based and not generalizable method to extract and compare answers.
% self-consistency is a rule-based and not generalizable method to extract and compare answers, while multi-agents (debate) has a generalizable answer extraction yet rule-based answer comparison way. 
However, self-agreement is an entirely generalizable method, whether extracting answers or comparing answers. 

\begin{table}[t]
  \centering
  \small
  \resizebox{1.0\linewidth}{!}{
  \begin{tabular}{@{}ll  @{\hspace{2pt}} cc}
      \toprule
      \bf Model & & \multicolumn{1}{c}{\textit{Arithmetic}} & \multicolumn{1}{c}{\textit{Commonsense}} \\
      \cmidrule(r){3-3}
      \cmidrule(r){4-4}
       & & \bf GSM8K & \bf CSQA \\\midrule
       \multicolumn{4}{c}{\textit{Both the type of questions and the answer format are known (\textbf{the third scenario})}} \\
       \midrule
       \multirow{3}{*}{Llama-2-70B-Chat} & Few-Shot CoT & 48.1 & 77.1 \\
       & Self-Consistency & 59.7 & 78.0 \\
       & Self-Agreement & 61.0 (\blue{+12.9}) & 77.7 (\blue{+0.6}) \\
       \midrule
       \multirow{3}{*}{GPT-4} & Few-Shot CoT & 90.8 & 86.9 \\
       & Self-Consistency & 93.3 & 88.1 \\
       & Self-Agreement & 93.9 (\blue{+3.1}) & 87.9 (\blue{+1.0}) \\
      \bottomrule
  \end{tabular}
  }
  \caption{
  Our performance gain (\blue{\textbf{+blue}}) is computed over few-shot CoT~\cite{wei2022chain}, which is our primary baseline.
  Our methods substantially increase the performance over~\citet{wei2022chain}.
  }
  \label{table:t25}
\end{table}
From the perspective of experiments, in order to promote the development of this field and test the general effectiveness of our method, we build \textit{an open-ended arithmetic dataset}, \textbf{GSM8K-Multi}, based on GSM8K~\cite{cobbe2021training}, where a reasoning path to a question has two different sub-answers consisting of the final answer. The reason for building the dataset in the arithmetic category is that GSM8K is a more complicated dataset than others (the dataset with the lowest baseline performance) to better measure the effectiveness of different methods. To build this dataset, we only select 500 examples randomly from GSM8K to save labor costs. We then hire crowd-sourced workers who
are Chinese-English bilingual speakers with enough mathematical knowledge to manually determine whether the math question can be rewritten into the type we need.\footnote{We follow two principles to rephrase the question. First, keep the difficulty of the original question as much as possible. Second, add a new problem under the original conditions and it does not conflict with the original problem.} If possible, directly rewrite the math question and ensure coherence between two different problems. If not, discard it. Table~\ref{table:t20} provides examples of original questions and rephrased questions of GSM8K. The corresponding solutions are also rewritten by them to guarantee the accuracy of answers. We finally obtain 362 examples as the test set of GSM8K-Multi.
% \begin{table}[!t]
%     \small
%     \centering
%     \resizebox{1.0\linewidth}{!}{
%     \begin{tabular}{ccccccc}
%     \toprule
%     \bf Dataset & GSM8K & MultiArith & SVAMP & CSQA & ARC-c & Letter\\
%     \midrule
%     \bf Truncation Ratio (\%) & \bf 71.5 & 15.0 & 31.5 & 62.0 & 65.0 & 0.0\\
%     \bottomrule
%     \end{tabular}
%     }
%     \caption{Truncation ratio (\%) of the sentences generated by the first stage of zero-shot CoT in the first scenario.}
%     \label{table:t13}
% \end{table}
The entire dataset will be released soon. Thus, to validate the statement that self-agreement still can achieve substantially better performance over baselines, 
% while almost all previous ensemble-optimization work can not
we follow the same experimental settings used in the first scenario and conduct experiments on the GSM8K-Multi dataset. As shown in Table~\ref{table:t21}, overall speaking, self-agreement achieves better performance over all strong baselines on GSM8K-Multi, demonstrating that it is entirely generalizable method, whether extracting answers or comparing answers.\footnote{We hire the same crowd-sourced workers to manually evaluate the accuracy of the answers. We follow two principles for the fair and strict evaluation. First, only the accuracy of the final answer instead of the reasoning path is evaluated to save labor costs. 
% Second, if you answer only one of the problems correctly, you will be given 0.5 points, and if you answer all problems correctly, you will be given 1 point.
Second, the solution is correct only if both sub-answers consisting of the final answer are correct.} 
% While multi-agents (debate) performs even worse than zero-shot CoT as we expected. To better understand why multi-agents (debate) fails, we manually examined model generated all agents' outputs by GPT-3.5-turbo for GSM8K-Multi. We find there are two key reasons. First, it is difficult for multi-agents (debate) to extract the final answer consisting of two sub-answers. Second, determining whether two strings represent the same final answer is not applicable for it. For instance, ``Brad gets 20 dollars and Jeff gets 80 dollars'' and ``Jeff gets 80 dollars and Brad gets 20 dollars'' represent two different answers when employing multi-agents (debate) on GSM8K-Multi. Instead, self-agreement is entirely generalizable method, whether extracting answers or comparing answers.

In the future, we plan to design an automated evaluation method to save labor costs. As far as we know, there are some work to investigate how to use ChatGPT or GPT-4 as an automated evaluation metric for various NLP tasks~\cite{wang2023chatgpt}.

\begin{table}[t]
  \centering
  \small
  \resizebox{1.0\linewidth}{!}{
  \begin{tabular}{@{}ll  @{\hspace{2pt}} cc}
      \toprule
      \bf Model & & \multicolumn{2}{c}{\textit{Arithmetic}} \\
      \cmidrule(r){3-4}
       & & \bf GSM8K & \bf SVAMP \\\midrule
       \multicolumn{4}{c}{\textit{Both the type of questions and the answer format are known (\textbf{the third scenario})}} \\
       \midrule
       \multirow{6}{*}{GPT-3.5-turbo} & Few-Shot CoT & 70.0 & 82.0 \\
       & Self-Consistency & 80.3 & 85.9 \\
       & Self-Agreement & 82.4 (\blue{+12.4}) & 86.0 (\blue{+4.0}) \\
       \cmidrule(r){2-4}
       & Complex CoT & 82.8 & 80.3 \\
       & Self-Consistency & 88.5 & 87.3 \\
       & Self-Agreement & 88.3 (\blue{+5.5}) & 87.2 (\blue{+6.9}) \\
      \bottomrule
  \end{tabular}
  }
  \caption{
  Our performance gain (\blue{\textbf{+blue}}) is computed over few-shot CoT~\cite{wei2022chain} or Complex CoT~\cite{fu2022complexity}, which is our primary baseline.
  Our methods substantially increase the performance over~\citet{wei2022chain} and~\cite{fu2022complexity}.
  }
  \label{table:t26}
\end{table}
\subsection{Experiments with More Language Models}

To further demonstrate the general effectiveness of self-agreement with more language models, we conduct experiments on GSM8K and CSQA with Llama-2-70B-Chat and GPT-4 in the third scenario. The reason we choose GSM8K and CSQA is that they are more complicated than others (the dataset with the lowest baseline performance). As shown in Table~\ref{table:t25}, self-agreement consistently improves over the few-shot CoT, and the performance is generally comparable to the standard self-consistency, which self-agreement does not need answer parsing to perform the voting.

\subsection{Experiments with More CoT Prompting Methods}

To further demonstrate the general effectiveness of self-agreement on more CoT prompting methods, we reproduce one of the very popular CoT prompting work, i.e., Complex CoT~\cite{fu2022complexity}, and conduct experiments on GSM8K and SVAMP with GPT-3.5-turbo in the third scenario. The reason we choose GSM8K and SVAMP in arithmetic reasoning is that they are more complicated than others (the dataset with the lowest baseline performance). As shown in Table~\ref{table:t26}, self-agreement consistently improves over the Complex CoT, indicating that it is applicable and beneficial to various prompting methods.

\section{Discusstion with USC}
\label{sec:dis}

After completion of this work, we find that~\citet{chen2023universal} share almost the same idea with us, which is archived after our work. So, we would like to discuss the differences between our work and it. \textbf{Firstly}, our work was archived on 14 Nov 2023, however, they were 29 Nov 2023. So, it is clear and certain that we are the first to propose the real-world scenarios in ensemble-optimization studies, and design a simple ensemble-optimization method applying in almost all scenarios. \textbf{Secondly}, Our work studies ensemble-optimization in multi-step reasoning situations, i.e., \textbf{open-ended reasoning tasks}, however, USC is designed to solve \textbf{open-ended generation tasks}. This is our future work we have discussed in Appendix~\ref{sec:appdis}. In principle the idea of this method can be extended to any open-ended generation tasks, not just open-ended reasoning tasks if a good metric can be well defined to compare multiple generations, where the good metric or overall procedure can be achieved by prompting language model itself one more time. So, we are very happy to see that USC has implemented the idea we have discussed in Appendix~\ref{sec:appdis}. \textbf{Thirdly}, \textit{we are the first to explore a more realistic setting with significant application values in ensemble-optimization studies}, as shown in Figure~\ref{fig:fig2}. Furthermore, to prompt this field, we build an open-ended arithmetic dataset, GSM8K-Multi, based on GSM8K~\cite{cobbe2021training}, where a reasoning path to a question has two different sub-answers consisting of the final answer, resulting in a more free-form and complicated answer formats. However, they only test the effectiveness of their method on two public fix-ended reasoning tasks. \textbf{Fourthly}, the prompt used in this work is different from that used in USC. We design a more domain-specific prompt to select the optimal response, which can further boost USC. This is also demonstrated in~\citet{chen2023universal}. \textbf{Fifthly}, \textit{self-agreement achieves better performance than USC}, indicating that self-agreement has better expandability and can really approach the performance ceiling of self-consistency while USC cannot. To be specific, for self-agreement, sampling a higher number of reasoning paths leads to a consistently better performance, while the opposite is true for USC. So, we strongly question the statements in this paper, i.e., the performance is generally comparable to the standard self-consistency, which USC does not need answer parsing to perform the voting. It is unfair to compare with self-consistency when the number of sampled reasoning paths is 8 rather than 40 (experimental settings in self-consistency paper). As shown in Table~\ref{table:t12}, it can be seen that self-agreement matches the standard self-consistency performance without requiring the answer formats to be similar, however, the performance of USC is far inferior to self-consistency and self-agreement. Also, sampling a higher number of reasoning paths leads to a consistently better performance, while the opposite is true for USC. Therefore, we consider \textbf{there is an inescapable gap between performance and generalization}. We focus on how to solve open-ended reasoning tasks, not open-ended generation tasks. Because the ability of LLMs to challenging tasks (e.g., arithmetic, commonsense and symbolic reasoning) is often seen as a limitation, which is difficult to be overcome solely by scaling up the size of LLMs.

\section{Case Study}
\label{sec:acs}

We also provide examples generated by self-agreement for different reasoning tasks in the first scenario, when applied on GPT-3.5-turbo. Table~\ref{table:t10},~\ref{table:t15},~\ref{table:t16},~\ref{table:t17},~\ref{table:t18},~\ref{table:t19} represent corresponding reasoning tasks. We show self-agreement can extract answers from reasoning paths to different type of input questions, compare different answers and then generate the majority voted answer.

\section{Broader Impacts}
\label{sec:appdis}

This work investigates the critical role of language model itself in the ensemble-optimization studies. Self-agreement allows language model itself to extract final answers of multiple reasoning paths, count the votes of different answers, and then select the majority voted answer. The overall procedure can be achieved by \textit{prompting language model itself} one more time. 

Self-agreement first explores an interesting direction, i.e., open-ended reasoning tasks. It takes an important step forward in ensemble-optimization studies where most of them are focused on fix-ended reasoning tasks. However, in principle the idea of this method can be extended to any open-ended generation tasks, not just open-ended reasoning tasks if a good metric can be well defined to compare multiple generations, where the good metric or overall procedure can be achieved by \textit{prompting language model itself} one more time. 

% Our findings encourage the research community to focus on a deeper understanding of the role of language model itself in ensemble-optimization studies.

\begingroup
\begin{table*}
    \centering
    \resizebox{1.0\linewidth}{!}{
    \begin{tabular}{p{0.96\linewidth}}
         \toprule
         \vspace{-1mm}
         \textbf{Original Question:} Adrien's total salary was 30 percent higher than Lylah's. Four years later, his salary had increased, and he was earning 40\% more than what he was making four years ago. If Adrien's and Lylah's salary increased simultaneously, and Adrien earned \$40000 four years ago, \textcolor{green}{calculate the total salary the two were receiving four years later}? \\
         \textbf{Rephrased Question:} Adrien's total salary was 30 percent higher than Lylah's. Four years later, his salary had increased, and he was earning 40\% more than what he was making four years ago. If Adrien's and Lylah's salary increased simultaneously, and Adrien earned \$40000 four years ago, \textcolor{cyan}{please calculate the total salary and salary difference for both of them after four years}? \\
         \midrule
         \textbf{Original Question:} John buys 2 pairs of shoes for each of his 3 children. They cost \$60 each. \textcolor{green}{How much did he pay}? \\
         \textbf{Rephrased Question:} John buys 2 pairs of shoes for each of his 3 children. They cost \$60 each. \textcolor{cyan}{How much does each child's shoes cost}? \textcolor{cyan}{How much did John pay in total}? \\
         \bottomrule
    \end{tabular}
    }
    \caption{Examples of original questions and rephrased questions of GSM8K~\cite{cobbe2021training}.}
    \label{table:t20}
\end{table*}
\endgroup
\begingroup
\begin{table*}[t]
    \centering
    \resizebox{1.0\linewidth}{!}{
    \begin{tabular}{p{0.96\linewidth}}
        \toprule
        \underline{\textbf{\textsc{Original Prompt for Reasoning Tasks}}} \\
        \vspace{0mm}
        When making a majority vote on multiple corresponding solutions while solely relying on the final answer of each solution, follow these detailed steps: \\
        \vspace{-1mm}
        1.Understand the Question: Begin by thoroughly understanding the question, including its requirements, given conditions, and objectives. This is essential for evaluating each solution. \\
        \vspace{-1mm}
        2.Extract the final answer for each solution: Extract the final answer for each solution, and then organize them in a clear list for ease of comparison. \\
        \vspace{-1mm}
        3.Count the Votes: Tally the number of times each final answer appears to determine which one received the majority of votes. \\
        \vspace{-1mm}
        4.Select the Final Answer: The final answer that received the most votes is chosen as the majority choice. In the case of a tie, you can either choose one of the tied answers. \\
        \vspace{-1mm}
        Below is a question and several candidate solutions or one candidate solution. By considering all these steps above, you can make a majority vote on several corresponding solutions, and then write ``The majority voted answer is'' in the last line. The majority voted answer should be one of the final answer sets extracted from several candidate solutions, not a certain solution. \\
        \bottomrule
    \end{tabular}
    }
    \caption{
    Original prompt used in the \textit{ask one more time} stage of \textbf{self-agreement} method.
    }
    \label{table:t5}
\end{table*}
\endgroup

\begin{table*}[t]
\setlength{\tabcolsep}{5.5pt}
\centering
\resizebox{1.0\linewidth}{!}{
\begin{tabular}{c|c|l}
    \toprule
     {\bf Task} & Type & {\bf Prompt} \\
      \midrule
      \multirow{5}{*}{General}  & \multirow{2}{*}{Starting} & \emph{Can you solve the following problem? <Problem> Explain your reasoning. Your final answer should be }  \\
      & & \emph{in the form \textbackslash boxed\{\{answer\}\}, at the end of your response.} \\
      \cmidrule{2-3}
     & \multirow{3}{*}{Debate} & \emph{These are the solutions to the problem from other agents: <other agent responses> Using the solutions from } \\
     & & \emph{other agents as additional information, can you provide your answer to the problem? The original problem } \\
     & & \emph{is <Problem>. Your final answer should be in the form} \emph{ \textbackslash boxed\{\{answer\}\}, at the end of your response.} \\
    \bottomrule
\end{tabular}
}
\caption{General prompts are used for each task.}
\label{table:t8}
\end{table*}

\begingroup
\begin{table*}[t]
    \centering
    \resizebox{1.0\linewidth}{!}{
    \begin{tabular}{p{0.96\linewidth}}
        \toprule
        \underline{\textbf{\textsc{Modified Prompt for Reasoning Tasks}}} \\
        \vspace{0mm}
        When majority voting on multiple corresponding solutions and relying only on the final answer for each solution, follow the steps detailed below: \\
        \vspace{-1mm}
        1.Understanding the question: The first step is to thoroughly understand the question, including its requirements, given conditions and objectives. This is essential for evaluating each solution. \\
        \vspace{-1mm}
        2.Extract the final answer for each solution: Extract the final answer for each solution and then organize them in a clear list for easy comparison. \\
        \vspace{-1mm}
        3.Counting of votes: The number of times each final answer appears is counted to determine which answer receives the majority of votes. \\
        \vspace{-1mm}
        4.Choose the final answer: The final answer that receives the most votes is chosen as the majority choice. In case of a tie, you can choose one of the tied answers. \\
        \vspace{-1mm}
        Below is a question and several candidate solutions or one candidate solution. By considering all the steps above, you can take a majority vote on several corresponding solutions and then write ``The majority voted answer is'' in the last line. The majority voted answer should be one of the final answer sets extracted from several candidate solutions, not a certain solution. \\
        \bottomrule
    \end{tabular}
    }
    \caption{
    Modified prompt used in the \textit{ask one more time} stage of \textbf{self-agreement} method.
    }
    \label{table:t6}
\end{table*}
\endgroup

\begingroup
\begin{table*}[htb]
    \centering
    \resizebox{1.0\linewidth}{!}{
    \begin{tabular}{p{0.96\linewidth}}
        \toprule
        \textbf{Q:} There are 15 trees in the grove. Grove workers will plant trees in the grove today. After they are done, there will be 21 trees. How many trees did the grove workers plant today? \\
\vspace{-1mm}
\textbf{A:} \hl{There are 15 trees originally. Then there were 21 trees after some more were planted. So there must have been 21 - 15 = 6.} The answer is 6. \\
\vspace{0mm}
\textbf{Q:} If there are 3 cars in the parking lot and 2 more cars arrive, how many cars are in the parking lot? \\
\vspace{-1mm}
\textbf{A:} \hl{There are originally 3 cars. 2 more cars arrive. 3 + 2 = 5.} The answer is 5. \\
\vspace{0mm}
\textbf{Q:} Leah had 32 chocolates and her sister had 42. If they ate 35, how many pieces do they have left in total? \\
\vspace{-1mm}
\textbf{A:} \hl{Originally, Leah had 32 chocolates. Her sister had 42. So in total they had 32 + 42 = 74. After eating 35, they had 74 - 35 = 39.} The answer is 39. \\
\vspace{0mm}
\textbf{Q:} What do people use to absorb extra ink from a fountain pen? Answer Choices: (A) shirt pocket (B) calligrapher's hand (C) inkwell (D) desk drawer (E) blotter \\
\vspace{-1mm}
\textbf{A:} \hl{The answer must be an item that can absorb ink. Of the above choices, only blotters are used to absorb ink.} The answer is E. \\
\vspace{0mm}
\textbf{Q:} George wants to warm his hands quickly by rubbing them. Which skin surface will produce the most heat? Answer Choices: (A) dry palms. (B) wet palms. (C) palms covered with oil. (D) palms covered with lotion. \\
\vspace{-1mm}
\textbf{A:} \hl{Dry surfaces will more likely cause more friction via rubbing than other smoother surfaces, hence dry palms will produce the most heat.} The answer is A. \\
\vspace{0mm}
\textbf{Q:} Take the last letters of the words in "Elon Musk" and concatenate them. \\
\vspace{-1mm}
\textbf{A:} \hl{The last letter of ``Elon'' is ``n''. The last letter of ``Musk'' is ``k''. Concatenating them is ``nk''.} The answer is nk. \\
\bottomrule
    \end{tabular}
    }
    \caption{
    Few-shot exemplars for full chain of thought in the second scenario. This set of exemplars is used for all reasoning tasks.
    }
    \label{table:t7}
\end{table*}
\endgroup

\begingroup
\begin{table*}
    \centering
    \small
    \begin{tabular}{p{0.96\linewidth}}
        \toprule
        \vspace{-1mm}
         \textbf{Q:} Jame will turn 27 in 5 years.  In 8 years his cousin will be 5 years younger than twice his age. How many years separate the age of the two now? \\
         \midrule
         \multicolumn{1}{c}{\textbf{\textit{Round 1}}} \\
         \midrule
         \vspace{-1mm}
         % \textbf{Agent 1 (role):} Can you solve the following problem? Jame will turn 27 in 5 years.  In 8 years his cousin will be 5 years younger than twice his age. How many years separate the age of the two now? Explain your reasoning. Your final answer should be in the form \textbackslash boxed\{\{answer\}\}, at the end of your response. \\ 
         % \vspace{-1mm}
         % \textbf{Agent 2 (role):} Can you solve the following problem? Jame will turn 27 in 5 years.  In 8 years his cousin will be 5 years younger than twice his age. How many years separate the age of the two now? Explain your reasoning. Your final answer should be in the form \textbackslash boxed\{\{answer\}\}, at the end of your response. \\ 
         % \vspace{-1mm}
         % \textbf{Agent 3 (role):} Can you solve the following problem? Jame will turn 27 in 5 years.  In 8 years his cousin will be 5 years younger than twice his age. How many years separate the age of the two now? Explain your reasoning. Your final answer should be in the form \textbackslash boxed\{\{answer\}\}, at the end of your response. \\
         % \midrule
         % \vspace{-1mm}
         \textbf{Agent 1:} Sure, I'd be happy to help! Let's start by analyzing the information given: * Jame will turn 27 in 5 years. * In 8 years, his cousin will be 5 years younger than twice his age.  We can start by finding out how old Jame will be in 8 years:  Jame's age in 8 years = 27 + 8 = 35  Now, let's find out how old his cousin will be in 8 years:  Cousin's age in 8 years = Jame's age in 8 years - 5 = 35 - 5 = 30  Since his cousin will be 5 years younger than twice Jame's age, we can find out Jame's current age by subtracting 5 from twice his age:  Twice Jame's age = 2 x 27 = 54  Current age = 54 - 5 = 49  So, Jame is currently 49 years old, and his cousin is 30 years old. The difference between their ages is:  49 - 30 = 19  Therefore, the answer is:  \textbackslash boxed\{\textcolor{red}{19}\} This means that Jame is currently 19 years older than his cousin. \\
         \vspace{-1mm}
         \textbf{Agent 2:} Sure, I'd be happy to help! Let's start by analyzing the information we have:  1. Jame will turn 27 in 5 years. 2. In 8 years, his cousin will be 5 years younger than twice his age.  We can start by finding out how old Jame will be in 8 years:  Jame's age in 8 years = 27 + 8 = 35  Now, we can find out how old his cousin will be in 8 years:  Cousin's age in 8 years = Jame's age in 8 years - 5 = 35 - 5 = 30  So, the difference between Jame's age and his cousin's age in 8 years is:  35 - 30 = 5  Therefore, the age difference between Jame and his cousin now is 5 years.  The answer is: \textbackslash boxed\{\textcolor{red}{5}\}. \\ 
         \vspace{-1mm}
         \textbf{Agent 3:} Sure, I'd be happy to help! Let's start by analyzing the information we have:  * Jame will turn 27 in 5 years. * In 8 years, his cousin will be 5 years younger than twice his age.  We can start by finding out how old Jame's cousin will be in 8 years. Since Jame will be 27 in 5 years, his cousin will be 27 - 5 = 22 years old in 8 years.  Now, we know that Jame's cousin will be 5 years younger than twice his age, so we can set up the equation:  Cousin's age = 22 - 5  Since Jame is currently 22 years old, we can find out the difference in their ages by subtracting Jame's age from his cousin's age:  Difference in ages = Cousin's age - Jame's age = 22 - 22 = 0  Therefore, Jame and his cousin are currently the same age.  So, the answer to the problem is: \textbackslash boxed\{\textcolor{red}{0 years}\}  There is no difference in age between Jame and his cousin currently, and they will both be the same age in 5 years when Jame turns 27. \\
         \midrule
         \multicolumn{1}{c}{\textbf{\textit{Round 2}}} \\
         \midrule
         \vspace{-1mm}
        \textbf{Agent 1:} Sure! Based on the solutions provided by the other agents, we can start by analyzing the information we have:  * Jame will turn 27 in 5 years. * In 8 years, his cousin will be 5 years younger than twice his age.  One agent solution 1 suggests that the age difference between Jame and his cousin now is 5 years. However, the other agent solution 2 states that Jame and his cousin are currently the same age.  Let's assume that the age difference between Jame and his cousin now is 5 years, as suggested by agent solution 1. This means that Jame's cousin is currently 22 years old (since Jame is 22 years old now and will be 27 in 5 years).  Now, we can calculate the age difference between Jame and his cousin in 8 years, based on the information provided:  In 8 years, Jame will be 27 + 8 = 35 years old. His cousin will be 22 + 8 = 30 years old.  The age difference between Jame and his cousin in 8 years is:  35 - 30 = 5 years  Therefore, the age difference between Jame and his cousin now is 5 years, as suggested by agent solution 1.  However, if we assume that Jame and his cousin are currently the same age, as suggested by agent solution 2, then the age difference between them now is 0 years.  So, the answer to the problem could be either \textbackslash boxed\{5 years\} or \textbackslash boxed\{0 years\}, depending on which assumption we make about the current age difference between Jame and his cousin.  Therefore, our final answer is:  \textbackslash boxed\{\textcolor{red}{5 years or 0 years}\}, depending on the assumption made about the current age difference between Jame and his cousin. \\
         \textbf{Agent 2:} Sure, I'd be happy to help! Let's start by analyzing the information we have:  * Jame will turn 27 in 5 years. * In 8 years, his cousin will be 5 years younger than twice his age.  We can start by finding out how old Jame's cousin will be in 8 years. Since Jame will be 27 in 5 years, his cousin will be 27 - 5 = 22 years old in 8 years.  Now, we know that Jame's cousin will be 5 years younger than twice his age, so we can set up the equation:  Cousin's age = 22 - 5  Since Jame is currently 22 years old, we can find out the difference in their ages by subtracting Jame's age from his cousin's age:  Difference in ages = Cousin's age - Jame's age = 22 - 22 = 0  Therefore, Jame and his cousin are currently the same age.  So, the answer to the problem is:  \textbackslash boxed\{\textcolor{red}{0 years}\}  There is no difference in age between Jame and his cousin currently, and they will both be the same age in 5 years when Jame turns 27.  Using the solutions from other agents, we can see that one agent found the answer to be 19 years, while another agent found the answer to be 0 years. However, based on the information provided, the correct answer is 0 years, as Jame and his cousin are currently the same age. \\
         \textbf{Agent 3:} Sure, I'd be happy to help! Based on the information provided, we can start by finding out how old Jame will be in 8 years:  Jame's age in 8 years = 27 + 8 = 35  Now, we can find out how old his cousin will be in 8 years:  Cousin's age in 8 years = Jame's age in 8 years - 5 = 35 - 5 = 30  So, the difference between Jame's age and his cousin's age in 8 years is:  35 - 30 = 5  However, we are also given that Jame's cousin will be 5 years younger than twice Jame's age. We can use this information to find out Jame's current age:  Twice Jame's age = 2 x 27 = 54  Current age = 54 - 5 = 49  So, Jame is currently 49 years old, and his cousin is 30 years old. The difference between their ages is:  49 - 30 = 19  Therefore, the answer is:  \textbackslash boxed\{\textcolor{red}{19}\}  This means that Jame is currently 19 years older than his cousin. \\
         \bottomrule
    \end{tabular}
    \caption{An example of the detailed debating process of 3 agents, 2 rounds.}
    \label{table:t9}
\end{table*}
\endgroup

\begingroup
\begin{table*}
    \centering
    \resizebox{1.0\linewidth}{!}{
    \begin{tabular}{p{0.96\linewidth}}
         \toprule
         \vspace{-1mm}
         \textbf{Question:} The caretaker of the docks needs to buy some new line. He wants 3 feet of line for every foot of dock. Right now, there is 200 feet of dock, and he has 6 feet of new line. How many feet of line does he need to buy in total? \\
         \midrule
         \textbf{Greedy Decode:} First, we need to calculate how much line the caretaker needs for the current dock. We know that he needs 3 feet of line for every foot of dock, so: 3 feet of line per foot of dock x 200 feet of dock = 600 feet of line needed for the current dock \textcolor{red}{Next, we need to add the 6 feet of new line that the caretaker already has: 600 feet of line needed for the current dock + 6 feet of new line = 606 feet of line needed in total} Therefore, the caretaker needs to buy \textcolor{red}{606} feet of line in total. The answer is \textcolor{red}{606} feet of line. \\
         \midrule
         \vspace{-1mm}
         \textbf{Sampled Path 1:} 1. The caretaker needs 3 feet of line for every foot of dock, so for 200 feet of dock, he needs: 3 x 200 = 600 feet of line 2. He already has 6 feet of new line, so he still needs: \textcolor{blue}{600 - 6 = 594 feet of line 3}. Therefore, the caretaker needs to buy \textcolor{blue}{594} feet of line in total.\\
         \vspace{-1mm}
         \textbf{Sampled Path 2:} 1. We know that he wants 3 feet of line for every foot of dock. So, for 200 feet of dock, he would need: 3 feet of line x 200 feet of dock = 600 feet of line 2. However, he already has 6 feet of new line. So, we need to subtract that from the total he needs: \textcolor{blue}{600 feet of line - 6 feet of new line = 594 feet of line} Therefore, the caretaker needs to buy \textcolor{blue}{594} feet of line in total. \\
         \midrule
         \vspace{-1mm}
         \textbf{Self-Agreement:} The majority voted answer is: \textcolor{blue}{594} feet of line. \\
         \bottomrule
    \end{tabular}
    }
    \caption{GSM8K examples where self-agreement generates the majority voted answer based on diverse reasoning paths, on GPT-3.5-turbo.}
    \label{table:t10}
\end{table*}
\endgroup

\begingroup
\begin{table*}
    \centering
    \resizebox{1.0\linewidth}{!}{
    \begin{tabular}{p{0.96\linewidth}}
         \toprule
         \vspace{-1mm}
         \textbf{Question:} There were 16 friends playing a video game online when 7 players quit. If each player left had 8 lives, how many lives did they have total? \\
         \midrule
         \textbf{Greedy Decode:} Step 1: Find the number of players who quit the game.There were 16 friends playing the game online, and 7 players quit. So, the number of players who remained in the game is: 16 - 7 = 9 Step 2: Find the total number of lives the players who quit had. Each player who quit had 8 lives. So, the total number of lives the players who quit had is: \textcolor{red}{7 x 8 = 56} Therefore, the players who quit had a total of \textcolor{red}{56} lives. The answer is \textcolor{red}{56}. \\
         \midrule
         \vspace{-1mm}
         \textbf{Sampled Path 1:} Step 1: Determine how many players were left after 7 players quit. 16 - 7 = 9 Step 2: Multiply the number of players left by the number of lives each player had. 9 x 8 = 72 Therefore, the players who quit had a total of 56 lives (7 players x 8 lives each) and the players who remained had a total of \textcolor{blue}{72 lives (9 players x 8 lives each)}.\\
         \vspace{-1mm}
         \textbf{Sampled Path 2:} First, we need to find out how many players were left after 7 quit. 16 - 7 = 9 players were left. Next, we need to multiply the number of players left by the number of lives each player had before they quit. \textcolor{blue}{9 players x 8 lives = 72 lives total}. Therefore, the 7 players who quit had no impact on the total number of lives, and the remaining 9 players had a total of \textcolor{blue}{72} lives. \\
         \midrule
         \vspace{-1mm}
         \textbf{Self-Agreement:} The majority voted answer is: \textcolor{blue}{72} lives. \\
         \bottomrule
    \end{tabular}
    }
    \caption{MultiArith examples where self-agreement generates the majority voted answer based on diverse reasoning paths, on GPT-3.5-turbo.}
    \label{table:t15}
\end{table*}
\endgroup

\begingroup
\begin{table*}
    \centering
    \resizebox{1.0\linewidth}{!}{
    \begin{tabular}{p{0.96\linewidth}}
         \toprule
         \vspace{-1mm}
         \textbf{Question:} Marco and his dad went strawberry picking. Together they collected strawberries that weighed 24 pounds. On the way back Marco lost 9 pounds of strawberries. Marco's strawberries now weighed 3 pounds. How much did his dad's strawberries weigh? \\
         \midrule
         \textbf{Greedy Decode:} 1. We know that Marco and his dad collected strawberries that weighed 24 pounds. 2. Marco lost 9 pounds of strawberries on the way back, so he was left with 3 pounds. 3. To find out how much his dad's strawberries weighed, we can subtract Marco's 3 pounds from the original 24 pounds: \textcolor{red}{24 pounds - 3 pounds = 21 pounds} Therefore, Marco's dad's strawberries weighed \textcolor{red}{21} pounds. The answer is \textcolor{red}{21} pounds. \\
         \midrule
         \vspace{-1mm}
         \textbf{Sampled Path 1:} First, we need to find out how many pounds of strawberries Marco had before he lost 9 pounds. We know that they collected 24 pounds in total, so: 24 pounds - 9 pounds = 15 pounds So Marco had 15 pounds of strawberries before he lost 9 pounds. Now we know that Marco's strawberries weigh 3 pounds, so we can subtract that from the 15 pounds he had before: \textcolor{blue}{15 pounds - 3 pounds = 12 pounds} Therefore, Marco's dad must have had \textcolor{blue}{12} pounds of strawberries.\\
         \vspace{-1mm}
         \textbf{Sampled Path 2:} 1. We know that Marco and his dad collected strawberries that weighed 24 pounds. 2. Marco lost 9 pounds of strawberries on the way back, so he was left with 3 pounds. This means that his dad's strawberries weighed \textcolor{blue}{24 - 9 - 3 = 12 pounds}. 3. Therefore, Marco's dad's strawberries weighed \textcolor{blue}{12} pounds. \\
         \midrule
         \vspace{-1mm}
         \textbf{Self-Agreement:} The majority voted answer is: Marco's dad's strawberries weighed \textcolor{blue}{12} pounds. \\
         \bottomrule
    \end{tabular}
    }
    \caption{SVAMP examples where self-agreement generates the majority voted answer based on diverse reasoning paths, on GPT-3.5-turbo.}
    \label{table:t16}
\end{table*}
\endgroup

\begingroup
\begin{table*}
    \centering
    \resizebox{1.0\linewidth}{!}{
    \begin{tabular}{p{0.96\linewidth}}
         \toprule
         \vspace{-1mm}
         \textbf{Question:} John was my neighbor, it was easy to talk to him. He was never what? Answer Choices: (A) away (B) distant (C) remote person (D) bore (E) foe \\
         \midrule
         \textbf{Greedy Decode:} The sentence says that John was easy to talk to, which means he was approachable and friendly. So, we can eliminate options (B), (C), and (E) as they all suggest that John was not friendly or easy to talk to. Option (A) suggests that John was not always available, which is not relevant to the sentence. \textcolor{red}{Therefore, the correct answer is (D) bore, which means John was not boring or uninteresting to talk to.} The answer is \textcolor{red}{(D)} bore. \\
         \midrule
         \vspace{-1mm}
         \textbf{Sampled Path 1:} The sentence implies that John is someone who is easy to talk to. Therefore, we can eliminate answer choices (B), (C), and (E) because they all suggest that John is not approachable or friendly. Answer choice (A), ``away,'' doesn't make sense in this context because it doesn't describe John's personality or behavior. That leaves us with answer choice (D), ``bore.'' This also doesn't fit the context because the sentence implies that John is easy to talk to, which suggests that he is not boring. Therefore, the correct answer is \textcolor{blue}{(B) distant}. \\
         \vspace{-1mm}
         \textbf{Sampled Path 2:} The sentence states that John was easy to talk to, which implies that he was approachable. Therefore, we can eliminate options (C) and (E) since they describe someone who is not approachable. The sentence does not provide any information about John's availability or whether he was often absent, so we can eliminate option (A). The sentence also does not suggest that John was boring, so we can eliminate option (D). Therefore, the correct answer is \textcolor{blue}{(B) distant, which means John was not aloof or reserved}. \\
         \midrule
         \vspace{-1mm}
         \textbf{Self-Agreement:} The majority voted answer is \textcolor{blue}{(B)} distant. \\
         \bottomrule
    \end{tabular}
    }
    \caption{CommonseQA examples where self-agreement generates the majority voted answer based on diverse reasoning paths, on GPT-3.5-turbo.}
    \label{table:t17}
\end{table*}
\endgroup

\begingroup
\begin{table*}
    \centering
    \resizebox{1.0\linewidth}{!}{
    \begin{tabular}{p{0.96\linewidth}}
         \toprule
         \vspace{-1mm}
         \textbf{Question:} Which nonrenewable resource is used extensively in computers and electronics? Answer Choices: (A) oil (B) gas (C) gold (D) lead \\
         \midrule
         \textbf{Greedy Decode:} Computers and electronics require materials that can conduct electricity and withstand high temperatures. Oil and gas are not used in the production of computers and electronics. Gold is a good conductor of electricity, but it is a rare and expensive metal, so it is not used extensively in electronics. \textcolor{red}{Lead, on the other hand, is a common metal that is used extensively in electronics, especially in soldering}. Therefore, the correct answer is \textcolor{red}{(D)} lead. \\
         \midrule
         \vspace{-1mm}
         \textbf{Sampled Path 1:} Computers and electronics require materials that can conduct electricity and are good at storing information. Option (A) and (B) are fossil fuels and not components of computers and electronics. Option (D) is used in batteries and some electronic components, but it is not as extensively used as the answer option. Option (C) - \textcolor{blue}{gold - is used extensively in computers and electronics because it is an excellent conductor of electricity and does not corrode or tarnish over time}. Therefore, the correct answer is \textcolor{blue}{(C)} gold.\\
         \vspace{-1mm}
         \textbf{Sampled Path 2:} Computers and electronics require a variety of materials to function, including metals, plastics, and various components. One nonrenewable resource that is used extensively in computers and electronics is gold. \textcolor{blue}{Gold is an excellent conductor of electricity and is used in circuit boards, connectors, and other electronic components}. Therefore, the correct answer is \textcolor{blue}{(C)} gold. \\
         \midrule
         \vspace{-1mm}
         \textbf{Self-Agreement:} The majority voted answer is \textcolor{blue}{(C)} gold.\\
         \bottomrule
    \end{tabular}
    }
    \caption{ARC-c examples where self-agreement generates the majority voted answer based on diverse reasoning paths, on GPT-3.5-turbo.}
    \label{table:t18}
\end{table*}
\endgroup

\begingroup
\begin{table*}
    \centering
    \resizebox{1.0\linewidth}{!}{
    \begin{tabular}{p{0.96\linewidth}}
         \toprule
         \vspace{-1mm}
         \textbf{Question:} Take the last letters of each words in ``Maria Elena Dewayne Mj Elliott'' and concatenate them. \\
         \midrule
         \textbf{Greedy Decode:} 1. Maria - last letter is ``a'' 2. Elena - last letter is ``a'' 3. Dewayne - last letter is ``e'' 4. Mj - last letter is ``j'' 5. Elliott - last letter is ``t'' Concatenating these letters gives us ``\textcolor{red}{aaejt}''. The answer is ``\textcolor{red}{aaejt}''. \\
         \midrule
         \vspace{-1mm}
         \textbf{Sampled Path 1:} 1. The last letter of ``Maria'' is ``a''. 2. The last letter of ``Elena'' is ``a''. 3. The last letter of ``Dewayne'' is ``e''. 4. The last letter of ``Mj'' is ``j''. 5. The last letter of ``Elliott'' is ``t''. Therefore, the concatenated last letters of each word is ``\textcolor{blue}{aejt}''.\\
         \vspace{-1mm}
         \textbf{Sampled Path 2:} 1. The last letter of ``Maria'' is ``a''. 2. The last letter of ``Elena'' is ``a''. 3. The last letter of ``Dewayne'' is ``e''. 4. The last letter of ``Mj'' is ``j''. 5. The last letter of ``Elliott'' is ``t''. So, the concatenated result is: ``\textcolor{blue}{aejt}''. \\
         \midrule
         \vspace{-1mm}
         \textbf{Self-Agreement:} The majority voted answer is ``\textcolor{blue}{aejt}''. \\
         \bottomrule
    \end{tabular}
    }
    \caption{Last Letters examples where self-agreement generates the majority voted answer based on diverse reasoning paths, on GPT-3.5-turbo.}
    \label{table:t19}
\end{table*}
\endgroup

\end{document}